\documentclass[11pt,a4paper]{article}
\usepackage[hyperref]{emnlp2020}
\usepackage{times}
\usepackage{latexsym}

\usepackage{graphicx}

\usepackage{microtype}
\usepackage{multirow}
\usepackage{booktabs}
\usepackage{xspace}
\usepackage{caption}
\usepackage{subcaption}
\usepackage{enumitem}

\aclfinalcopy 
\def\aclpaperid{1263} 

\newcommand*{\yoruba}{Yor\`ub\'a\xspace}

\title{Transfer Learning and Distant Supervision for Multilingual Transformer Models: A Study on African Languages}

\author{Michael A. Hedderich\textsuperscript{1}, David I. Adelani\textsuperscript{1}, Dawei Zhu\textsuperscript{1}, Jesujoba Alabi\textsuperscript{1,2}, Udia Markus\textsuperscript{3} \\ \textbf{\& Dietrich Klakow\textsuperscript{1}}\\
\textsuperscript{1}Saarland University, Saarland Informatics Campus, Germany\\
\textsuperscript{2}DFKI GmBH, Saarbrücken, Germany
\textsuperscript{3}Nuhu Bamalli Polytechnic, Zaira, Nigeria\\
\texttt{\{mhedderich,didelani,dzhu,dietrich.klakow\}@lsv.uni-saarland.de} \\
\texttt{jesujoba\_oluwadara.alabi@dfki.de }
}

\date{}

\begin{document}
\maketitle
\begin{abstract}
Multilingual transformer models like mBERT and XLM-RoBERTa have obtained great improvements for many NLP tasks on a variety of languages. However, recent works also showed that results from high-resource languages could not be easily transferred to realistic, low-resource scenarios. In this work, we study trends in performance for different amounts of available resources for the three African languages Hausa, isiXhosa and \yoruba on both NER and topic classification. We show that in combination with transfer learning or distant supervision, these models can achieve with as little as 10 or 100 labeled sentences the same performance as baselines with much more supervised training data. However, we also find settings where this does not hold. Our discussions and additional experiments on assumptions such as time and hardware restrictions highlight challenges and opportunities in low-resource learning. 
\end{abstract}

\section{Introduction}

Deep learning techniques, including contextualized word embeddings based on transformers and pretrained on language modelling, have resulted in considerable improvements for many NLP tasks. However, they often require large amounts of labeled training data, and there is also growing evidence that transferring approaches from high to low-resource settings is not straightforward. In \cite{african/Loubser2020}, rule-based or linguistically motivated CRFs still outperform RNN-based methods on several tasks for South African languages. For pretraining approaches where labeled data exists in a high-resource language, and the information is transferred to a low-resource language, \citet{data/Xtreme20} find a significant gap between performance on English and the cross-lingually transferred models. In a recent study, \citet{lowresource/Lauscher2020FromZTH} find that the transfer for multilingual transformer models is less effective for resource-lean settings and distant languages. A popular technique to obtain labeled data quickly and cheaply is distant and weak supervision. \citet{lowresource/kann20weakly} recently inspected POS classifiers trained on weak supervision. They found that in contrast to scenarios with simulated low-resource settings of high-resource languages, in truly low-resource settings this is still a difficult problem. These findings also highlight the importance of aiming for realistic experiments when studying low-resource scenarios. 

In this work, we analyse multilingual transformer models, namely mBERT \cite{models/BERT, models/mBERT} and XLM-RoBERTa \cite{models/RoBERTa}. We evaluate both sequence and token classification tasks in the form of news title topic classification and named entity recognition (NER). A variety of approaches have been proposed to improve performance in low-resource settings. In this work, we study (i) transfer learning from a high-resource language and (ii) distant supervision. We selected these as they are two of the most popular techniques in the recent literature and are rather independent of a specific model architecture. Both need auxiliary data. For transfer learning, this is labeled data in a high-resource language, and for distant supervision, this is expert insight and a mechanism to (semi-)automatically generate labels. We see them, therefore, as orthogonal and depending on the scenario and the data availability, either one or the other approach might be applicable.

Our study is performed on three, linguistically different African languages: Hausa, isiXhosa and \yoruba. These represent languages with millions of users and active use of digital infrastructure, but with only very limited support for NLP technologies. For this aim, we also collected three new datasets that are made publicly available alongside the code and additional material.\footnote{\url{https://github.com/uds-lsv/transfer-distant-transformer-african}}

We show both challenges and opportunities when working with multilingual transformer models evaluating trends for different levels of resource scarcity. The paper is structured into the following questions we are interested in:
\vspace{-0.25cm}
\begin{itemize}[leftmargin=*]
    \item How do more complex transformer models compare to established RNNs?\vspace{-0.25cm}
    \item How can transfer-learning be used effectively?\vspace{-0.25cm}
    \item Is distant supervision helpful?\vspace{-0.25cm}
    \item  What assumptions do we have to consider when targeting a realistic treatment of low-resource scenarios?
\end{itemize} 

\section{Languages and Datasets}
In this work, we evaluate on three African languages, namely Hausa, isiXhosa and \yoruba. Hausa is from the Afro-Asiatic family while isiXhosa and \yoruba belong to different branches of the large Niger-Congo family. Hausa and \yoruba are the second and third most spoken languages in Africa, and isiXhosa is recognized as one of the official languages in South Africa and Zimbabwe. \yoruba has been part of the unlabeled training data for the mBERT multilingual, contextual word embeddings. Texts in Hausa and isiXhosa have been part of the XLM-RoBERTa training.

The three languages have few or no labeled datasets online for popular NLP tasks like named entity recognition (NER) and topic classification. We use the NER dataset by \citet{african/Eiselen16NERData} for isiXhosa and the one by \citet{alabi-LREC} for \yoruba. We collected and manually annotated a NER dataset for Hausa and news title topic classification datasets for Hausa and \yoruba. Table~\ref{tab:dataset_info} gives a summary of the datasets. More information about the languages, the datasets and their creation process can be found in the Appendix.

\begin{table}[t]
    \centering
    \footnotesize
    \begin{tabular}{p{25mm}cp{22mm}}
    \toprule
     \textbf{Dataset Name} &  \textbf{Data Source} &  \textbf{Full Train/ Val/ Test} \textbf{sentences} \\ \addlinespace[0.2em]
    \midrule
      Hausa NER* & VOA Hausa &  1,014 / 145 / 291  \\ \addlinespace[0.2em]
      Hausa Topic Class.* & VOA Hausa  & 2,045 / 290 /582  \\ \addlinespace[0.2em]
      isiXhosa NER~\cite{african/Eiselen16NERData}  & SADiLaR
      & 5,138 / 608 / 537  \\ \addlinespace[0.2em]
      \yoruba NER~\cite{alabi-LREC} & GlobalVoices
      & 816 / 116 / 236  \\ \addlinespace[0.2em]
      \yoruba~Topic~Class.* & BBC Yoruba & 1,340 / 189 / 379  \\ \addlinespace[0.2em]
      \bottomrule
    \end{tabular}
  \caption{Datasets Summary. *Created for this work. \label{tab:dataset_info}}
\end{table}

\section{Experimental Settings}
To evaluate different amounts of resource-availability, we use subsets of the training data with increasing sizes from ten to the maximally available number of sentences. All the models are trained on their corresponding language-model pretraining. Except if specified otherwise, the models are not fine-tuned on any other task-specific, labeled data from other languages. We report mean F1-score on the test sets over ten repetitions with standard error on the error bars. Additional experimental details are given in the following sections and the Appendix. The code is made publicly available online as well as a table with the scores for all the runs.

\section{Comparing to RNNs}
\label{sec:rnn}
\begin{figure}[t]
\centering
     \begin{subfigure}[b]{\columnwidth}
        \centering
        \includegraphics[height=5cm]{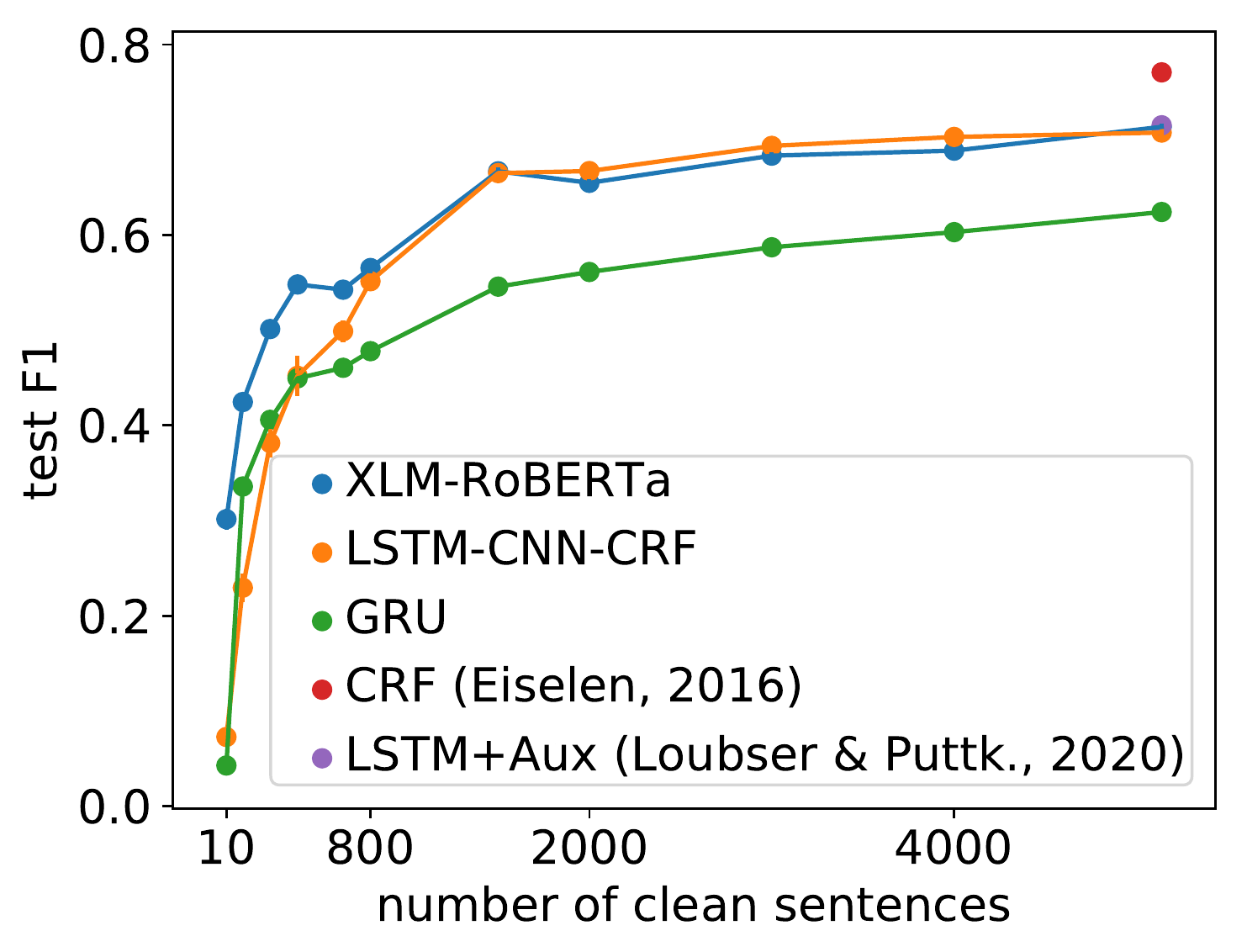}
        \vspace{-0.2cm}
        \caption{NER isiXhosa}
    \end{subfigure}
    \begin{subfigure}[b]{\columnwidth}
        \centering
        \includegraphics[height=5cm]{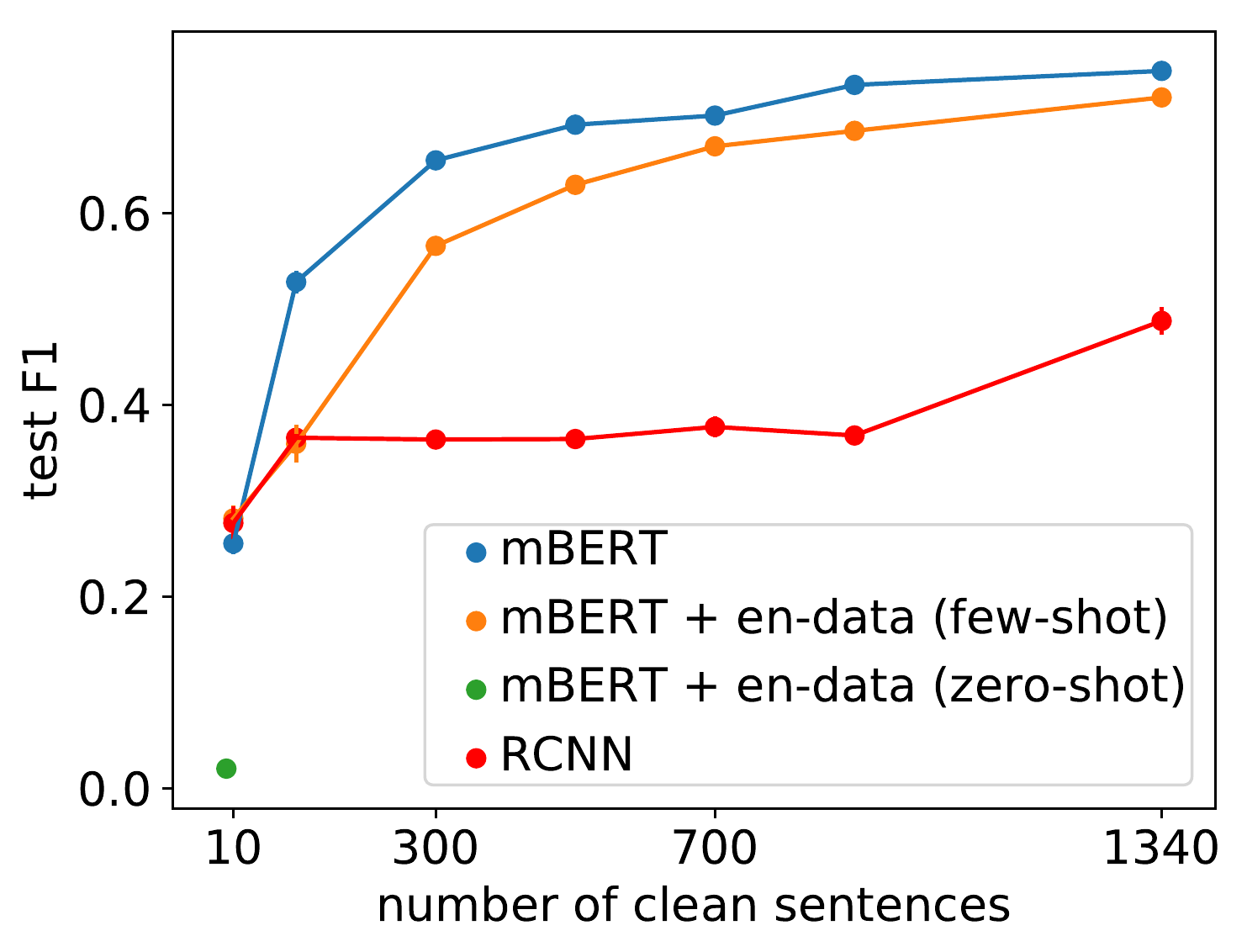}
        \vspace{-0.2cm}
        \caption{Topic Class. \yoruba}
    \end{subfigure}
    \begin{subfigure}[b]{\columnwidth}
    \centering
        \includegraphics[height=5cm]{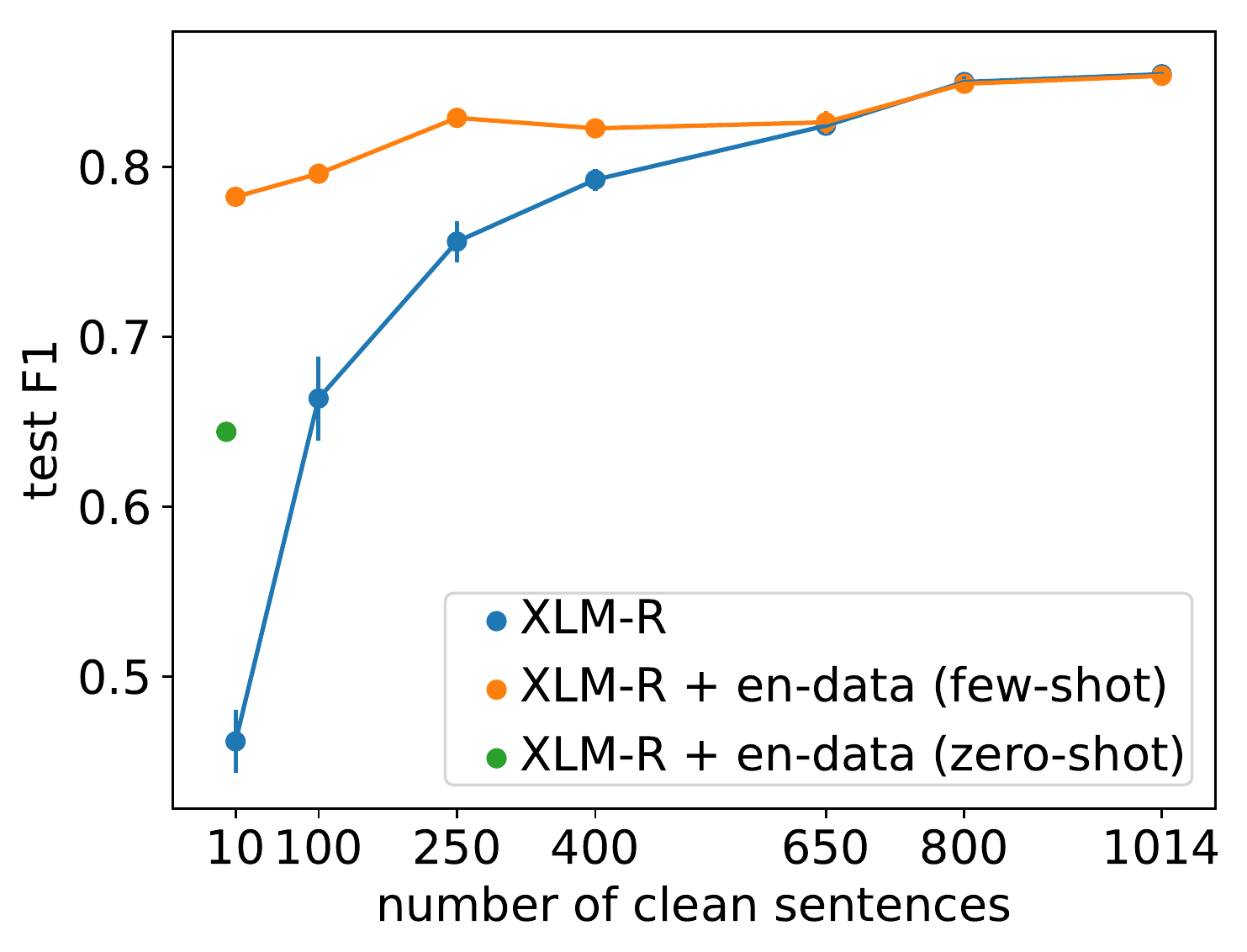}
        \vspace{-0.2cm}
        \caption{Transfer Learn NER Hausa}
    \end{subfigure}
     \vspace{-0.2cm}
        \caption{Comparing to RNNs (a+b) and using transfer learning (b+c). Additional plots in the Appendix.\label{fig:rnn_transfer}}
        \vspace{-0.4cm}
\end{figure}

\citet{african/Loubser2020} showed that models with comparatively few parameters, like CRFs, can still outperform more complex, neural RNNs models for several task and low-resource language combinations. This motivates the question whether model complexity is an issue for these low-resource NLP models. We compare to simple GRU based \cite{GRU} models as well as the popular (non-transformer) combination of LSTM-CNN-CRF \cite{ma-hovy-2016-end} for NER and to the RCNN architecture \cite{lai2015recurrent} for topic classification. For these models, we use pre-trained, non-contextual word embeddings trained for the specific language. Figures \ref{fig:rnn_transfer}a+b show that an increase in model complexity is not an issue in these experiments. For Hausa and \yoruba and for the low resource settings for isiXhosa, BERT and XLM-RoBERTa actually outperform the other baselines, possibly due to the larger amounts of background knowledge through the language model pre-training. For larger amounts of task-specific training data, the LSTM-CNN-CRF and the transformer models perform similarly. One should note that for isiXhosa, the linguistically motivated CRF \cite{african/Eiselen16NERData} still outperforms all approaches on the full dataset.

\section{Transfer Learning}
The mBERT and XLM-RoBERTa models are trained with tasks that can be obtained from unlabeled text, like masked language modelling. Additionally, the multilingual models can be fine-tuned on task-specific, supervised data but from a different, high-resource language. There is evidence that the multilingual transformer models can learn parallel concepts across languages \cite{pires-etal-2019-multilingual,wu-dredze-2019-beto,data/Xtreme20}. This allows to then apply or evaluate the model directly without having been fine-tuned on any labeled data in the target language (zero-shot) or on only a small amount of labeled data in the target language (few-shot).

For NER, we pre-train on the English CoNLL03 NER dataset \cite{data/CoNLL03}. For topic classification, the models are pre-trained on the English AG News corpus \cite{data/ag-news}. The texts in the high-resource English and the low-resource Hausa and \yoruba target datasets share the same domain (news texts). One issue that is visible in these experiments is the discrepancy between classes. While some classes like ``Politics'' are shared, the topic classification datasets also have language- and location-specific classes like ``Nigeria'' and ``Africa'' which are not part of the English fine-tuning dataset. In our experiments, we use the intersection of labels for NER (excluding DATE and MISC for Hausa and \yoruba) and the union of labels for topic classification. 

The results in Figure \ref{fig:rnn_transfer}c and in the Appendix confirm the benefits of fine-tuning on high-resource languages already shown in past research. They show, however, also the large gains in performance that can be obtained by training on a minimal number of target instances. While the zero-shot setting in \cite{data/Xtreme20} is interesting from a methodological perspective, using a small training set for the target language seems much more beneficial for a practical application. In our experiments, we get - with only ten labeled sentences - an improvement of at least 10 points in the F1-score for a shared label set on NER. For topic classification (Figure \ref{fig:rnn_transfer}b) the transfer learning is not beneficial, which might be due to the mismatch in the label sets.
\begin{figure*}[t]
\centering

\end{figure*}

\section{Distant Supervision}

Distant and weak supervision are popular techniques when labeled data is lacking. It allows a domain expert to insert their knowledge without having to label every instance manually. The expert can, e.g. create a set of rules that are then used to label the data automatically \cite{RatnerSnorkel20} or information from an external knowledge source can be used \cite{rijhwani2020soft}. This kind of (semi-) automatic supervision tends to contain more errors which can hurt the performance of classifiers (see e.g. \cite{fang-cohn-2016-learning}). To avoid this, it can be combined with label noise handling techniques. This pipeline has been shown to be effective for several NLP tasks \cite{lange-etal-2019-feature, paul-etal-2019-handling, wang-etal-2019-learning-noisy, chen-etal-2019-uncover, mayhew-etal-2019-named}, however, mostly for RNN based approaches. As we have seen in Section \ref{sec:rnn} that these have a lower baseline performance, we are interested in whether distant supervision is still useful for the better performing transformer models. Several of the past works evaluated their approach only on high-resource languages or simulated low-resource scenarios. We are, therefore, also interested in how the distant supervision performs for the actual resource-lean African languages we study.

To create the distant supervision,  native speakers with a background in NLP were asked to write labeling rules. For the NER labels PER, ORG and LOC, we match the tokens against lists of entity names. These were extracted from the corresponding categories from Wikidata. For the DATE label, the insight is used that date expressions are usually preceded by date keywords in \yoruba, as reported by \citet{adelani2020distant}. We find similar patterns in Hausa like ``\textit{ranar}''(day), ``\textit{watan}'' (month), and ``\textit{shekarar}''(year). For example, \textit{``18th of May, 2019''} in Hausa translates to \textit{``ranar 18 ga watan Mayu, shekarar 2019''}. The annotation rules are based on these keywords and further heuristics. Directly applying this distant supervision on the NER test sets results in an F1-score of $54\%$ and $62\%$ on Hausa and \yoruba, respectively.

For the topic classification task, the distant supervision rules are based on a dictionary of words relating to each of the classes. To induce the dictionaries, we collected terms related to different classes from web sources. For example, for the ``Sport'' label, names of sportspeople and sport-related organizations were collected and similarly for the ``Africa'' label, names of countries, their capitals and major cities and their politicians. To label a news headline, the intersection between each class-dictionary and the text was computed, and a class was selected with a majority voting scheme. We obtain an F1-score of $49\%$ and  $55\%$ on the Hausa and \yoruba test set respectively when applying the distant supervision directly to the topic classification test sets. Additional details on the distant supervision are given in the Appendix.

For label noise handling we use the confusion matrix approach for NER by \citet{noise/Hedderich18Confusion}, marked as \textit{cm} in the plots. Additionally, we propose to combine it with the smoothing concept by \citet{noise/LvSmoothing20}.

The Figures \ref{fig:distant_assumptions}a and in the Appendix show that when only a small amount of manually labeled data is available, distant supervision can be a helpful addition. E.g. for the NER task in \yoruba, combining distant supervision and noise handling with 100 labeled sentences achieves similar performance to using 400 manually labeled sentences. For label noise handling, combining the confusion matrix with the smoothing approach might be beneficial because the estimated confusion matrix is flawed when only small amounts of labeled data are given. When more manually labeled data is available, the noisy annotations lose their benefit and can become harmful to performance. Improved noise-handling techniques might be able to mitigate this.

\begin{figure}[h]
\centering
    \begin{subfigure}[b]{\columnwidth}
        \centering
        \includegraphics[height=5.2cm]{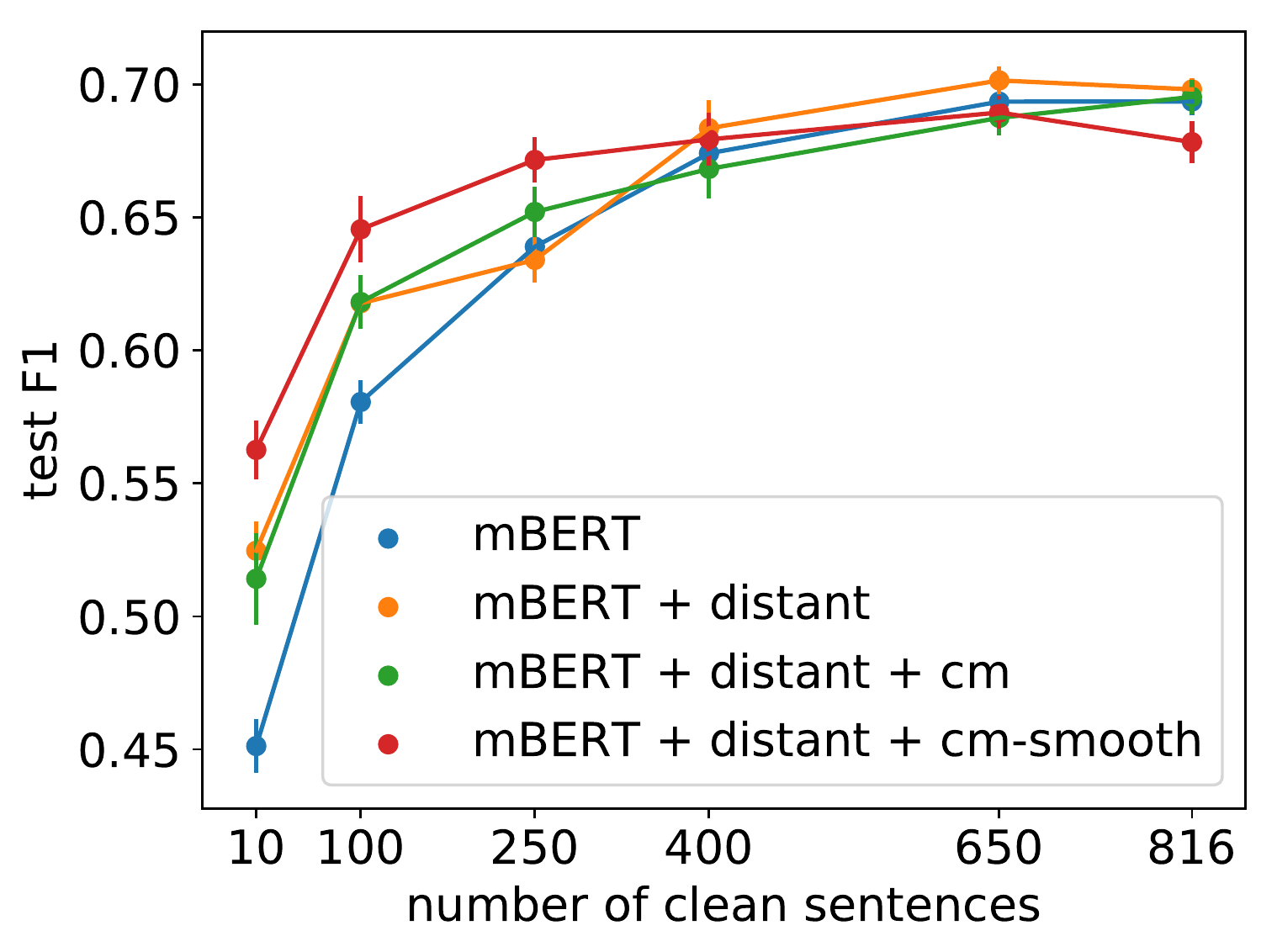}
        \vspace{-0.1cm}
        \caption{Distant Supervision NER \yoruba}
        \vspace{0.1cm}
    \end{subfigure}
    \begin{subfigure}[t]{\columnwidth}
        \centering
        \includegraphics[height=5.2cm]{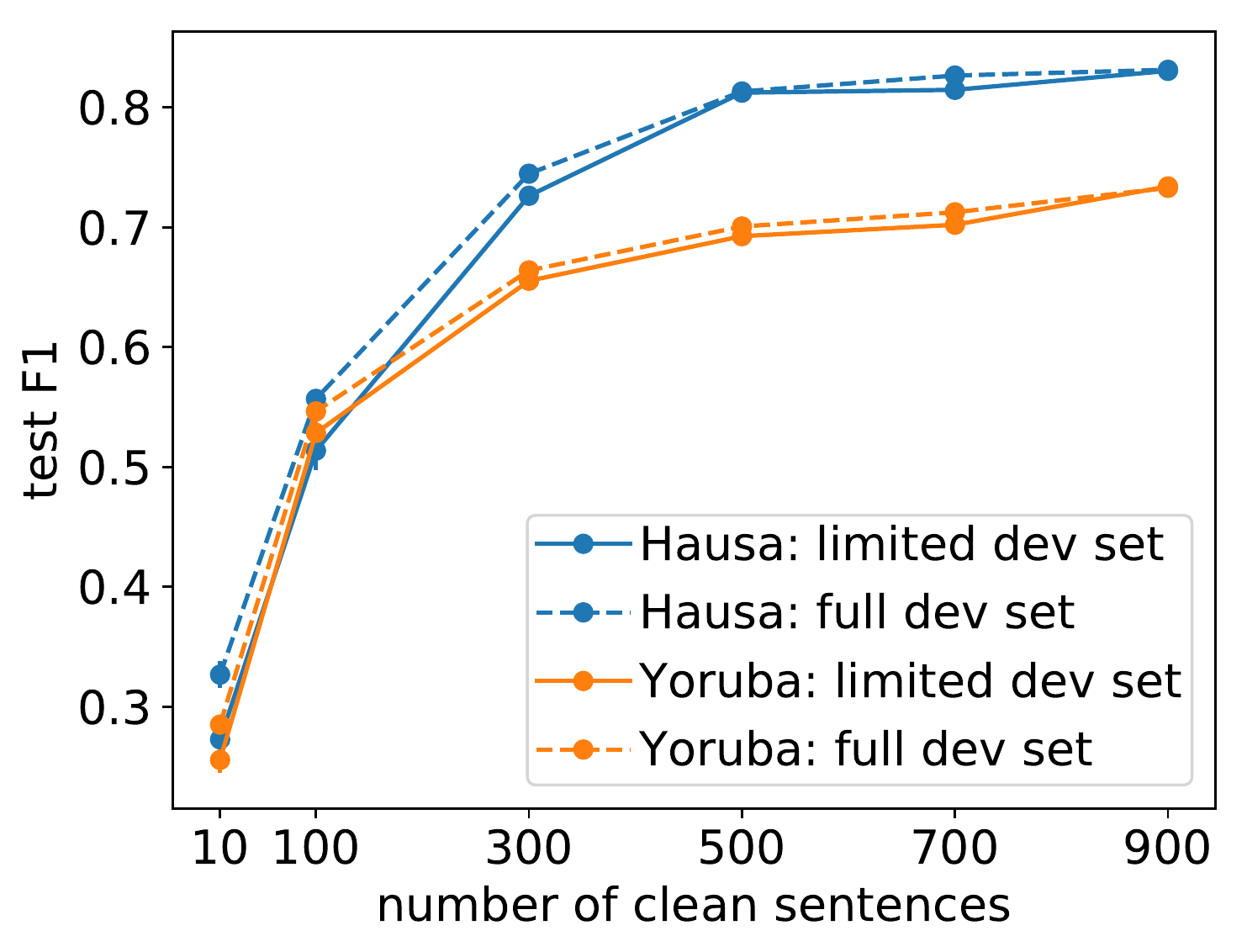}
        \vspace{-0.1cm}
        \caption{Development Set Topic Class.}
        \vspace{0.1cm}
    \end{subfigure}
     \begin{subfigure}[t]{\columnwidth}
        \centering
        \includegraphics[height=5.2cm]{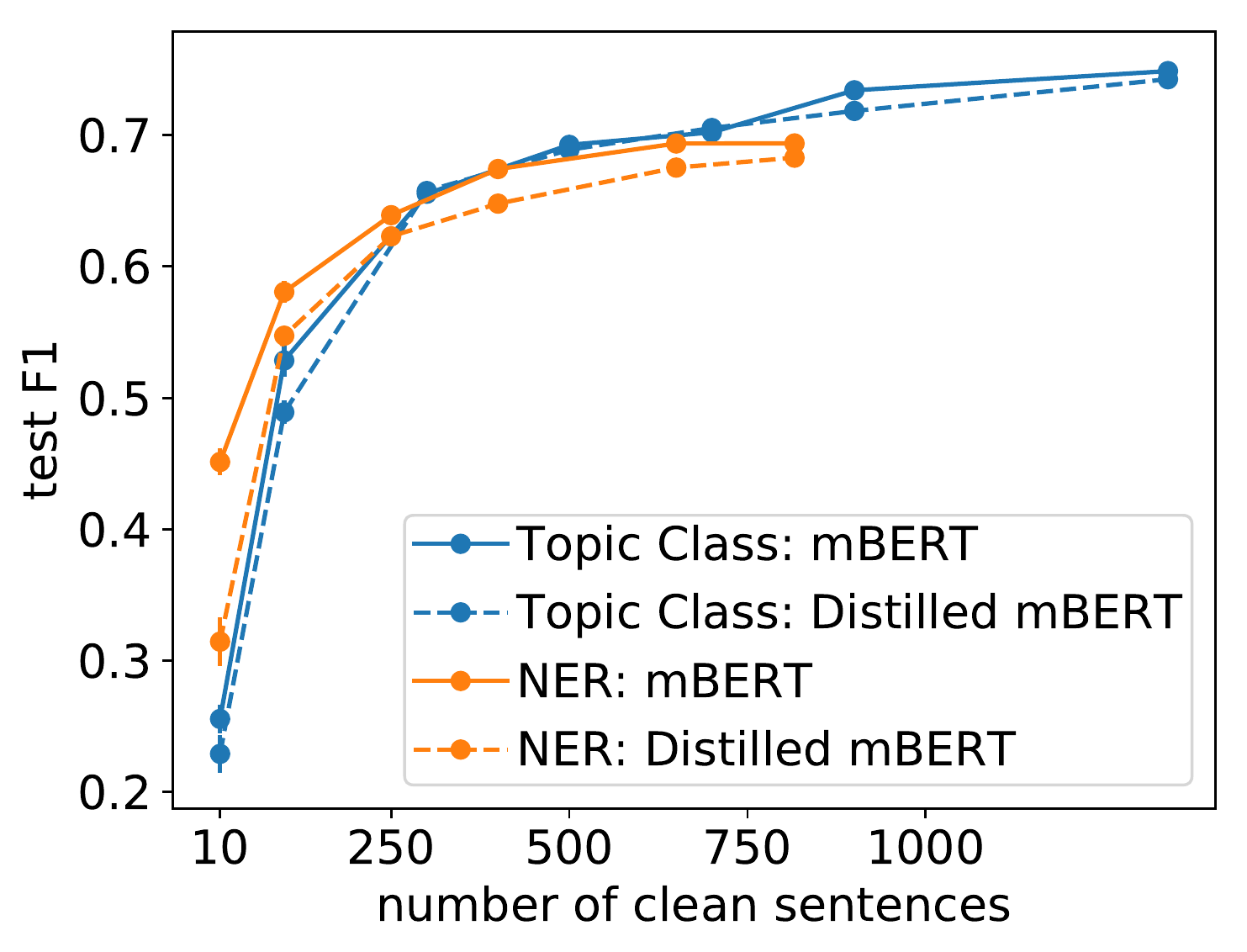}
        \vspace{-0.1cm}
        \caption{DistilBERT on \yoruba}
    \end{subfigure}
    \vspace{-0.1cm}
        \caption{Distant supervision and model variations. Additional plots in the Appendix.\label{fig:distant_assumptions}}
\end{figure}

\section{Questioning Assumptions}

In this section, we want to discuss certain assumptions taken by us and previous work in low-resource learning to see if these hold and what challenges and opportunities they could bring for future work.

\subsection{Development Set}
\citet{lowresource/kann19dev} criticized that research on low-resource often assumes the existence of a development set. Addressing this, we perform hyperparameter optimization on high-resource English data. For early-stopping (to avoid overfitting), the authors experiment with obtaining an early-stop-epoch from the average of several other languages. To avoid this multi-language set-up and the need to obtain labeled data for multiple languages, we suggest using instead a development set downsized by the same factor as the training data. This approach keeps the ratio between training and development set giving the development set a reasonable size to obtain in a low-resource setting. For the setting with ten labeled sentences for training, the same amount was used for the dev set. The results in Figure \ref{fig:distant_assumptions}b and in the Appendix show that this has only a small effect on the training performance.

\subsection{Hardware Resources}

While the multilingual transformer models show impressive improvements over the RNN baselines, they also require more hardware resources. The LSTM-CNN-CRF model, e.g. has ca. 5M parameters compared to mBERT's over 150M parameters. The computing capabilities for training and deploying such models might not always be given in low-resource scenarios. Through personal conversations with researchers from African countries, we found that this can be an issue. There are approaches to reduce model size while keeping a similar performance quality, e.g. the 25\% smaller DistilBERT \cite{sanh2019distilbert}. Figure \ref{fig:distant_assumptions}c shows that this performs indeed similar in many cases but that there is a significant drop in performance for NER when only few training sentences are available.

\subsection{Annotation Time}
In \cite{data/Xtreme20} and \cite{lowresource/kann20weakly}, it is assumed that no labeled training data is available for the target language. In the previous sections, we showed that even with ten labeled target sentences, reasonable model quality can be achieved. For our annotation efforts, we measured on average 1 minute per annotator per sentence for NER and 6 seconds per sentence for topic classification. We, therefore, think that it is reasonable to assume the availability of small amounts of labeled data. Especially, as we would argue that it is beneficial to have a native speaker or language expert involved when developing a model for a specific language.

For distant supervision, a trade-off arises given these annotation times. While extracting named entities from knowledge bases requires minimal manual effort assuming a set-up system, manual crafting rules took 30 minutes for the DATE label and 2.5 hours for each topic classification dataset. When reporting results for distant supervision, the performance benefits should therefore also be compared against manual annotation in the same time frame.

\section{Conclusions}
In this work, we evaluated transfer learning and distant supervision on multilingual transformer models, studying realistic low-resource settings for African languages. We show that even with a small amount of labeled data, reasonable performance can be achieved. We hope that our new datasets and our reflections on assumptions in low-resource settings help to foster future research in this area.

\section*{Acknowledgments}
We would like to thank Toyin Aina and Emmanuel Olawuyi for their support in annotating the Hausa data. We also thank Thomas Niesler for his helpful insights as well as Alexander Blatt and the anonymous reviewers for their feedback. Funded by the Deutsche Forschungsgemeinschaft (DFG, German Research Foundation) – Project-ID 232722074 – SFB 1102, the EU-funded Horizon 2020 projects ROXANNE under grant number 833635 and COMPRISE under grant agreement No. 3081705.
\bibliographystyle{acl_natbib}
\bibliography{emnlp2020}

\begin{thebibliography}{40}
\expandafter\ifx\csname natexlab\endcsname\relax\def\natexlab#1{#1}\fi

\bibitem[{Abdulmumin and Galadanci(2019)}]{Abdulmumin_2019}
Idris Abdulmumin and Bashir~Shehu Galadanci. 2019.
\newblock \href {https://doi.org/10.1109/nigeriacomputconf45974.2019.8949674}
  {hauwe: Hausa words embedding for natural language processing}.
\newblock \emph{2019 2nd International Conference of the IEEE Nigeria Computer
  Chapter (NigeriaComputConf)}.

\bibitem[{Adelani et~al.(2020)Adelani, Hedderich, Zhu, van~den Berg, and
  Klakow}]{adelani2020distant}
David~Ifeoluwa Adelani, Michael~A. Hedderich, Dawei Zhu, Esther van~den Berg,
  and Dietrich Klakow. 2020.
\newblock \href {http://arxiv.org/abs/2003.08370} {Distant supervision and
  noisy label learning for low resource named entity recognition: A study on
  hausa and yorùbá}.

\bibitem[{Alabi et~al.(2020)Alabi, Amponsah-Kaakyire, Adelani, and
  Espana-Bonet}]{alabi-LREC}
Jesujoba Alabi, Kwabena Amponsah-Kaakyire, David Adelani, and Cristina
  Espana-Bonet. 2020.
\newblock \href {https://www.aclweb.org/anthology/2020.lrec-1.335} {Massive vs.
  {C}urated {W}ord {E}mbeddings for {L}ow-{R}esourced {L}anguages. {T}he {C}ase
  of {Yor{\`u}b{\'a}} and {Twi}}.
\newblock In \emph{Proceedings of The 12th Language Resources and Evaluation
  Conference}, pages 2747--2755, Marseille, France. European Language Resources
  Association.

\bibitem[{Chen et~al.(2019)Chen, Zhang, Mao, Guo, and
  Xu}]{chen-etal-2019-uncover}
Junfan Chen, Richong Zhang, Yongyi Mao, Hongyu Guo, and Jie Xu. 2019.
\newblock \href {https://doi.org/10.18653/v1/D19-1031} {Uncover the
  ground-truth relations in distant supervision: A neural
  expectation-maximization framework}.
\newblock In \emph{Proceedings of the 2019 Conference on Empirical Methods in
  Natural Language Processing and the 9th International Joint Conference on
  Natural Language Processing (EMNLP-IJCNLP)}, pages 326--336, Hong Kong,
  China. Association for Computational Linguistics.

\bibitem[{Chernodub et~al.(2019)Chernodub, Oliynyk, Heidenreich, Bondarenko,
  Hagen, Biemann, and Panchenko}]{chernodub2019targer}
Artem Chernodub, Oleksiy Oliynyk, Philipp Heidenreich, Alexander Bondarenko,
  Matthias Hagen, Chris Biemann, and Alexander Panchenko. 2019.
\newblock Targer: Neural argument mining at your fingertips.
\newblock In \emph{Proceedings of the 57th Annual Meeting of the Association of
  Computational Linguistics (ACL'2019)}, Florence, Italy.

\bibitem[{Cho et~al.(2014)Cho, van Merrienboer, G{\"{u}}l{\c{c}}ehre, Bahdanau,
  Bougares, Schwenk, and Bengio}]{GRU}
Kyunghyun Cho, Bart van Merrienboer, {\c{C}}aglar G{\"{u}}l{\c{c}}ehre, Dzmitry
  Bahdanau, Fethi Bougares, Holger Schwenk, and Yoshua Bengio. 2014.
\newblock \href {https://doi.org/10.3115/v1/d14-1179} {Learning phrase
  representations using {RNN} encoder-decoder for statistical machine
  translation}.
\newblock In \emph{Proceedings of the 2014 Conference on Empirical Methods in
  Natural Language Processing, {EMNLP} 2014, October 25-29, 2014, Doha, Qatar,
  {A} meeting of SIGDAT, a Special Interest Group of the {ACL}}, pages
  1724--1734. {ACL}.

\bibitem[{Conneau et~al.(2019)Conneau, Khandelwal, Goyal, Chaudhary, Wenzek,
  Guzmán, Grave, Ott, Zettlemoyer, and Stoyanov}]{models/RoBERTa}
Alexis Conneau, Kartikay Khandelwal, Naman Goyal, Vishrav Chaudhary, Guillaume
  Wenzek, Francisco Guzmán, Edouard Grave, Myle Ott, Luke Zettlemoyer, and
  Veselin Stoyanov. 2019.
\newblock \href {http://arxiv.org/abs/1911.02116} {Unsupervised cross-lingual
  representation learning at scale}.

\bibitem[{Devlin(2019)}]{models/mBERT}
Jacob Devlin. 2019.
\newblock \href
  {https://github.com/google-research/bert/blob/cc7051dc592802f501e8a6f71f8fb3cf9de95dc9/multilingual.md}
  {{mBERT README} file}.

\bibitem[{Devlin et~al.(2019)Devlin, Chang, Lee, and Toutanova}]{models/BERT}
Jacob Devlin, Ming-Wei Chang, Kenton Lee, and Kristina Toutanova. 2019.
\newblock \href {https://doi.org/10.18653/v1/N19-1423} {{BERT}: Pre-training of
  deep bidirectional transformers for language understanding}.
\newblock In \emph{Proceedings of the 2019 Conference of the North {A}merican
  Chapter of the Association for Computational Linguistics: Human Language
  Technologies, Volume 1 (Long and Short Papers)}, pages 4171--4186,
  Minneapolis, Minnesota. Association for Computational Linguistics.

\bibitem[{Eberhard et~al.(2019)Eberhard, Simons, and (eds.)}]{Ethnologue2019}
David~M. Eberhard, Gary~F. Simons, and Charles D.~Fennig (eds.). 2019.
\newblock \href {http://www.ethnologue.com} {Ethnologue: Languages of the
  world. twenty-second edition.}

\bibitem[{Eiselen(2016)}]{african/Eiselen16NERData}
Roald Eiselen. 2016.
\newblock \href {https://www.aclweb.org/anthology/L16-1533} {Government domain
  named entity recognition for south {A}frican languages}.
\newblock In \emph{Proceedings of the Tenth International Conference on
  Language Resources and Evaluation ({LREC}'16)}, pages 3344--3348,
  Portoro{\v{z}}, Slovenia. European Language Resources Association (ELRA).

\bibitem[{Eiselen and Puttkammer(2014)}]{african/Eiselen14}
Roald Eiselen and Martin~J. Puttkammer. 2014.
\newblock \href
  {http://www.lrec-conf.org/proceedings/lrec2014/summaries/1151.html}
  {Developing text resources for ten south african languages}.
\newblock In \emph{Proceedings of the Ninth International Conference on
  Language Resources and Evaluation, {LREC} 2014, Reykjavik, Iceland, May
  26-31, 2014}, pages 3698--3703. European Language Resources Association
  {(ELRA)}.

\bibitem[{Fang and Cohn(2016)}]{fang-cohn-2016-learning}
Meng Fang and Trevor Cohn. 2016.
\newblock \href {https://doi.org/10.18653/v1/K16-1018} {Learning when to trust
  distant supervision: An application to low-resource {POS} tagging using
  cross-lingual projection}.
\newblock In \emph{Proceedings of The 20th {SIGNLL} Conference on Computational
  Natural Language Learning}, pages 178--186, Berlin, Germany. Association for
  Computational Linguistics.

\bibitem[{Hedderich and Klakow(2018)}]{noise/Hedderich18Confusion}
Michael~A. Hedderich and Dietrich Klakow. 2018.
\newblock \href {https://doi.org/10.18653/v1/W18-3402} {Training a neural
  network in a low-resource setting on automatically annotated noisy data}.
\newblock In \emph{Proceedings of the Workshop on Deep Learning Approaches for
  Low-Resource NLP, DeepLo@ACL 2018, Melbourne, Australia, July 19, 2018},
  pages 12--18. Association for Computational Linguistics.

\bibitem[{Hu et~al.(2020)Hu, Ruder, Siddhant, Neubig, Firat, and
  Johnson}]{data/Xtreme20}
Junjie Hu, Sebastian Ruder, Aditya Siddhant, Graham Neubig, Orhan Firat, and
  Melvin Johnson. 2020.
\newblock \href {http://arxiv.org/abs/2003.11080} {Xtreme: A massively
  multilingual multi-task benchmark for evaluating cross-lingual
  generalization}.

\bibitem[{Kann et~al.(2019)Kann, Cho, and Bowman}]{lowresource/kann19dev}
Katharina Kann, Kyunghyun Cho, and Samuel~R. Bowman. 2019.
\newblock \href {https://doi.org/10.18653/v1/D19-1329} {Towards realistic
  practices in low-resource natural language processing: The development set}.
\newblock In \emph{Proceedings of the 2019 Conference on Empirical Methods in
  Natural Language Processing and the 9th International Joint Conference on
  Natural Language Processing (EMNLP-IJCNLP)}, pages 3342--3349, Hong Kong,
  China. Association for Computational Linguistics.

\bibitem[{Kann et~al.(2020)Kann, Lacroix, and
  Søgaard}]{lowresource/kann20weakly}
Katharina Kann, Ophélie Lacroix, and Anders Søgaard. 2020.
\newblock \href {http://arxiv.org/abs/2004.13305} {Weakly supervised pos
  taggers perform poorly on truly low-resource languages}.

\bibitem[{Lai et~al.(2015)Lai, Xu, Liu, and Zhao}]{lai2015recurrent}
Siwei Lai, Liheng Xu, Kang Liu, and Jun Zhao. 2015.
\newblock Recurrent convolutional neural networks for text classification.
\newblock In \emph{Twenty-ninth AAAI conference on artificial intelligence}.

\bibitem[{Lange et~al.(2019)Lange, Hedderich, and
  Klakow}]{lange-etal-2019-feature}
Lukas Lange, Michael~A. Hedderich, and Dietrich Klakow. 2019.
\newblock \href {https://doi.org/10.18653/v1/D19-1362} {Feature-dependent
  confusion matrices for low-resource {NER} labeling with noisy labels}.
\newblock In \emph{Proceedings of the 2019 Conference on Empirical Methods in
  Natural Language Processing and the 9th International Joint Conference on
  Natural Language Processing (EMNLP-IJCNLP)}, pages 3554--3559, Hong Kong,
  China. Association for Computational Linguistics.

\bibitem[{Lauscher et~al.(2020)Lauscher, Ravishankar, Vulic, and
  Glavas}]{lowresource/Lauscher2020FromZTH}
Anne Lauscher, Vinit Ravishankar, Ivan Vulic, and Goran Glavas. 2020.
\newblock From zero to hero: On the limitations of zero-shot cross-lingual
  transfer with multilingual transformers.
\newblock \emph{ArXiv}, abs/2005.00633.

\bibitem[{Loubser and Puttkammer(2020{\natexlab{a}})}]{african/Loubser2020}
Melinda Loubser and Martin~J. Puttkammer. 2020{\natexlab{a}}.
\newblock Viability of neural networks for core technologies for
  resource-scarce languages.
\newblock \emph{Information}, 11:41.

\bibitem[{Loubser and Puttkammer(2020{\natexlab{b}})}]{Loubser2020ViabilityON}
Melinda Loubser and Martin~J. Puttkammer. 2020{\natexlab{b}}.
\newblock Viability of neural networks for core technologies for
  resource-scarce languages.
\newblock \emph{Information}, 11:41.

\bibitem[{Luo et~al.(2017)Luo, Feng, Wang, Zhu, Huang, Yan, and
  Zhao}]{luo-etal-2017-learning-noise}
Bingfeng Luo, Yansong Feng, Zheng Wang, Zhanxing Zhu, Songfang Huang, Rui Yan,
  and Dongyan Zhao. 2017.
\newblock \href {https://doi.org/10.18653/v1/P17-1040} {Learning with noise:
  Enhance distantly supervised relation extraction with dynamic transition
  matrix}.
\newblock In \emph{Proceedings of the 55th Annual Meeting of the Association
  for Computational Linguistics (Volume 1: Long Papers)}, pages 430--439,
  Vancouver, Canada. Association for Computational Linguistics.

\bibitem[{Lv et~al.(2020)Lv, Wu, and Xia}]{noise/LvSmoothing20}
Xianbin Lv, Dongxian Wu, and Shu{-}Tao Xia. 2020.
\newblock \href {http://arxiv.org/abs/2003.11904} {Matrix smoothing: {A}
  regularization for {DNN} with transition matrix under noisy labels}.
\newblock \emph{CoRR}, abs/2003.11904.

\bibitem[{Ma and Hovy(2016)}]{ma-hovy-2016-end}
Xuezhe Ma and Eduard Hovy. 2016.
\newblock \href {https://doi.org/10.18653/v1/P16-1101} {End-to-end sequence
  labeling via bi-directional {LSTM}-{CNN}s-{CRF}}.
\newblock In \emph{Proceedings of the 54th Annual Meeting of the Association
  for Computational Linguistics (Volume 1: Long Papers)}, pages 1064--1074,
  Berlin, Germany. Association for Computational Linguistics.

\bibitem[{Mayhew et~al.(2019)Mayhew, Chaturvedi, Tsai, and
  Roth}]{mayhew-etal-2019-named}
Stephen Mayhew, Snigdha Chaturvedi, Chen-Tse Tsai, and Dan Roth. 2019.
\newblock \href {https://doi.org/10.18653/v1/K19-1060} {Named entity
  recognition with partially annotated training data}.
\newblock In \emph{Proceedings of the 23rd Conference on Computational Natural
  Language Learning (CoNLL)}, pages 645--655, Hong Kong, China. Association for
  Computational Linguistics.

\bibitem[{Nivre et~al.(2020)Nivre, de~Marneffe, Ginter, Hajič, Manning,
  Pyysalo, Schuster, Tyers, and Zeman}]{nivre2020universal}
Joakim Nivre, Marie-Catherine de~Marneffe, Filip Ginter, Jan Hajič,
  Christopher~D. Manning, Sampo Pyysalo, Sebastian Schuster, Francis Tyers, and
  Daniel Zeman. 2020.
\newblock \href {http://arxiv.org/abs/2004.10643} {Universal dependencies v2:
  An evergrowing multilingual treebank collection}.

\bibitem[{Pan et~al.(2017)Pan, Zhang, May, Nothman, Knight, and
  Ji}]{pan-etal-2017-cross}
Xiaoman Pan, Boliang Zhang, Jonathan May, Joel Nothman, Kevin Knight, and Heng
  Ji. 2017.
\newblock \href {https://doi.org/10.18653/v1/P17-1178} {Cross-lingual name
  tagging and linking for 282 languages}.
\newblock In \emph{Proceedings of the 55th Annual Meeting of the Association
  for Computational Linguistics (Volume 1: Long Papers)}, pages 1946--1958,
  Vancouver, Canada. Association for Computational Linguistics.

\bibitem[{Paul et~al.(2019)Paul, Singh, Hedderich, and
  Klakow}]{paul-etal-2019-handling}
Debjit Paul, Mittul Singh, Michael~A. Hedderich, and Dietrich Klakow. 2019.
\newblock \href {https://doi.org/10.18653/v1/N19-3005} {Handling noisy labels
  for robustly learning from self-training data for low-resource sequence
  labeling}.
\newblock In \emph{Proceedings of the 2019 Conference of the North {A}merican
  Chapter of the Association for Computational Linguistics: Student Research
  Workshop}, pages 29--34, Minneapolis, Minnesota. Association for
  Computational Linguistics.

\bibitem[{Pires et~al.(2019)Pires, Schlinger, and
  Garrette}]{pires-etal-2019-multilingual}
Telmo Pires, Eva Schlinger, and Dan Garrette. 2019.
\newblock \href {https://doi.org/10.18653/v1/P19-1493} {How multilingual is
  multilingual {BERT}?}
\newblock In \emph{Proceedings of the 57th Annual Meeting of the Association
  for Computational Linguistics}, pages 4996--5001, Florence, Italy.
  Association for Computational Linguistics.

\bibitem[{Ratner et~al.(2020)Ratner, Bach, Ehrenberg, Fries, Wu, and
  R{\'{e}}}]{RatnerSnorkel20}
Alexander Ratner, Stephen~H. Bach, Henry~R. Ehrenberg, Jason~A. Fries, Sen Wu,
  and Christopher R{\'{e}}. 2020.
\newblock \href {https://doi.org/10.1007/s00778-019-00552-1} {Snorkel: rapid
  training data creation with weak supervision}.
\newblock \emph{{VLDB} J.}, 29(2):709--730.

\bibitem[{Rijhwani et~al.(2020)Rijhwani, Zhou, Neubig, and
  Carbonell}]{rijhwani2020soft}
Shruti Rijhwani, Shuyan Zhou, Graham Neubig, and Jaime Carbonell. 2020.
\newblock \href {http://arxiv.org/abs/2005.01866} {Soft gazetteers for
  low-resource named entity recognition}.

\bibitem[{Sanh et~al.(2019)Sanh, Debut, Chaumond, and
  Wolf}]{sanh2019distilbert}
Victor Sanh, Lysandre Debut, Julien Chaumond, and Thomas Wolf. 2019.
\newblock \href {http://arxiv.org/abs/1910.01108} {Distilbert, a distilled
  version of bert: smaller, faster, cheaper and lighter}.

\bibitem[{Strassel and Tracey(2016)}]{data/lorelei2016}
Stephanie Strassel and Jennifer Tracey. 2016.
\newblock \href {https://www.aclweb.org/anthology/L16-1521} {{LORELEI} language
  packs: Data, tools, and resources for technology development in low resource
  languages}.
\newblock In \emph{Proceedings of the Tenth International Conference on
  Language Resources and Evaluation ({LREC}'16)}, pages 3273--3280,
  Portoro{\v{z}}, Slovenia. European Language Resources Association (ELRA).

\bibitem[{Tjong Kim~Sang and De~Meulder(2003)}]{data/CoNLL03}
Erik~F. Tjong Kim~Sang and Fien De~Meulder. 2003.
\newblock \href {https://www.aclweb.org/anthology/W03-0419} {Introduction to
  the {C}o{NLL}-2003 shared task: Language-independent named entity
  recognition}.
\newblock In \emph{Proceedings of the Seventh Conference on Natural Language
  Learning at {HLT}-{NAACL} 2003}, pages 142--147.

\bibitem[{Tracey et~al.(2019)Tracey, Strassel, Bies, Song, Arrigo, Griffitt,
  Delgado, Graff, Kulick, Mott, and Kuster}]{data/lorelei2019}
Jennifer Tracey, Stephanie Strassel, Ann Bies, Zhiyi Song, Michael Arrigo, Kira
  Griffitt, Dana Delgado, Dave Graff, Seth Kulick, Justin Mott, and Neil
  Kuster. 2019.
\newblock \href {https://www.aclweb.org/anthology/W19-6808} {Corpus building
  for low resource languages in the {DARPA} {LORELEI} program}.
\newblock In \emph{Proceedings of the 2nd Workshop on Technologies for MT of
  Low Resource Languages}, pages 48--55, Dublin, Ireland. European Association
  for Machine Translation.

\bibitem[{Wang et~al.(2019)Wang, Liu, Li, Yang, and
  Li}]{wang-etal-2019-learning-noisy}
Hao Wang, Bing Liu, Chaozhuo Li, Yan Yang, and Tianrui Li. 2019.
\newblock \href {https://doi.org/10.18653/v1/D19-1655} {Learning with noisy
  labels for sentence-level sentiment classification}.
\newblock In \emph{Proceedings of the 2019 Conference on Empirical Methods in
  Natural Language Processing and the 9th International Joint Conference on
  Natural Language Processing (EMNLP-IJCNLP)}, pages 6286--6292, Hong Kong,
  China. Association for Computational Linguistics.

\bibitem[{Wolf et~al.(2019)Wolf, Debut, Sanh, Chaumond, Delangue, Moi, Cistac,
  Rault, Louf, Funtowicz, and Brew}]{wolf2019huggingfaces}
Thomas Wolf, Lysandre Debut, Victor Sanh, Julien Chaumond, Clement Delangue,
  Anthony Moi, Pierric Cistac, Tim Rault, Rémi Louf, Morgan Funtowicz, and
  Jamie Brew. 2019.
\newblock \href {http://arxiv.org/abs/1910.03771} {Huggingface's transformers:
  State-of-the-art natural language processing}.

\bibitem[{Wu and Dredze(2019)}]{wu-dredze-2019-beto}
Shijie Wu and Mark Dredze. 2019.
\newblock \href {https://doi.org/10.18653/v1/D19-1077} {Beto, bentz, becas: The
  surprising cross-lingual effectiveness of {BERT}}.
\newblock In \emph{Proceedings of the 2019 Conference on Empirical Methods in
  Natural Language Processing and the 9th International Joint Conference on
  Natural Language Processing (EMNLP-IJCNLP)}, pages 833--844, Hong Kong,
  China. Association for Computational Linguistics.

\bibitem[{Zhang et~al.(2015)Zhang, Zhao, and LeCun}]{data/ag-news}
Xiang Zhang, Junbo Zhao, and Yann LeCun. 2015.
\newblock \href
  {http://papers.nips.cc/paper/5782-character-level-convolutional-networks-for-text-classification.pdf}
  {Character-level convolutional networks for text classification}.
\newblock In C.~Cortes, N.~D. Lawrence, D.~D. Lee, M.~Sugiyama, and R.~Garnett,
  editors, \emph{Advances in Neural Information Processing Systems 28}, pages
  649--657. Curran Associates, Inc.

\end{thebibliography}

\appendix

\section{Languages}

In this work, we consider three languages: Hausa, isiXhosa and \yoruba.
These languages are from two language families: Niger-Congo and Afro-Asiatic, according to Ethnologue \cite{Ethnologue2019}, where the Niger-Congo family has over $20\%$ of the world languages.  

The Hausa language is native to the northern part of Nigeria and the southern part of the Republic of Niger with more than 45 million native speakers \cite{Ethnologue2019}. It is the second most spoken language in Africa after Swahili. Hausa is a tonal language, but this is not marked in written text. The language is written in a modified Latin alphabet. 

\yoruba, on the other hand, is native to south-western Nigeria and the Republic of Benin. It has over 35 million native speakers~\cite{Ethnologue2019} and is the third most spoken language in Africa. \yoruba is a tonal language with three tones: low, middle and high. These tones are represented by the grave (``\textbackslash
''), optional macron (``$-$'') and acute (``/'') accents respectively. The tones are represented in written texts along with a modified Latin alphabet.

Lastly, we consider isiXhosa, a Bantu language that is native to South Africa and also recognized as one of the official languages in South Africa and Zimbabwe. It is spoken by over 8 million native speakers~\cite{Ethnologue2019}. isiXhosa is a tonal language, but the tones are not marked in written text. The text is written with the Latin alphabet.

\citet{lowresource/kann20weakly} used as an indicator for a low-resource language the availability of data in the Universal Dependency project \cite{nivre2020universal}. The languages we study suit their indicator having less than 10k (\yoruba) or no data (Hausa, isiXhosa) at the time of writing.

\section{Datasets}

\subsection{Existing Datasets}
The WikiAnn corpus \cite{pan-etal-2017-cross} provides NER datasets for 282 languages available on Wikipedia. These are, however, only silver-standard annotations and for Hausa and isiXhosa less than 4k and 1k tokens respectively are provided. The LORELEI project announced the release of NER datasets for several African languages via LDC \cite{data/lorelei2016, data/lorelei2019} but have not yet done so for Hausa and \yoruba at the time of writing.

\citet{african/Eiselen14} and \citet{african/Eiselen16NERData} created NLP datasets for South African languages. We use the latter's NER dataset for isiXhosa. For the \yoruba NER dataset \cite{alabi-LREC}, we use the authors' split into training, dev and test set of the cased version of their data.\footnote{\url{https://github.com/ajesujoba/YorubaTwi-Embedding/tree/master/Yoruba/Yor\%C3\%B9b\%C3\%A1-NER}} For the isiXhosa dataset\footnote{\url{https://repo.sadilar.org/handle/20.500.12185/312}}, we use an 80\%/10\%/10\% split following the instructions in \cite{Loubser2020ViabilityON}. The split is based on token-count, splitting only after the end of the sentence (information obtained through personal conversation with the authors). For the fine-tuning of the zero- and few-shot models, the standard CoNLL03 NER \cite{data/CoNLL03} and AG News \cite{data/ag-news} datasets are used with their existing splits.

\subsection{New Datasets}

\subsubsection{Hausa NER}
For the Hausa NER annotation, we collected 250 articles from VOA Hausa\footnote{\url{https://www.voahausa.com}}, 50 articles each from the five pre-defined categories of the news website. The categories are Najeriya (Nigeria), Afirka (Africa), Amurka (USA), Sauran Duniya (the rest of the world) and Kiwon Lafiya (Health). We removed articles with less than 50 tokens which results in 188 news articles (over 37K tokens). We asked two volunteers who are native Hausa speakers to annotate the corpus separately. Each volunteer was supervised by someone with experience in NER annotation. Following the named entity annotation in \yoruba by \citet{alabi-LREC}, we annotated PER, ORG, LOC and DATE (dates and times) for Hausa. The annotation was based on the MUC-6 Named Entity Task Definition guide.\footnote{\url{https://cs.nyu.edu/faculty/grishman/NEtask20.book_1.html}} Comparing the annotations of the volunteers, we observe a conflict for 1302 tokens (out of 4838 tokens) excluding the non-entity words (i.e. words with 'O' labels). One of the annotators was better in annotating DATE, while the other was better in annotating ORG especially for multi-word expressions of entities. We resolved all the conflicts after discussion with one of the volunteers. The split of annotated data of the Yoruba and Hausa NER data is 70\%/10\%/20\% for training, validation and test sentences.

\subsubsection{Hausa and \yoruba Text classification}
For the topic classification datasets, news titles were collected from VOA Hausa and the BBC Yoruba news website\footnote{\url{https://www.bbc.com/yoruba}}. Two native speakers of the language annotated each dataset. We categorized the \yoruba news headlines into 7 categories, namely ``Nigeria'', ``Africa'', ``World'', ``Entertainment'', ``Health'', ``Sport'', ``Politics''. Similarly, we annotated 5 (of the 7) categories for Hausa news headlines, excluding ``Sport'' and ``Entertainment'' as there was only a limited number of examples. The ``Politics'' category in the annotation is only for Nigerian political news headlines. Comparing the two annotators, there was a conflict rate of 7.5\% for Hausa and 5.8\% for \yoruba. The total number of news titles after resolving conflicts was 2,917 for Hausa and 1,908 for \yoruba. 

\section{Word Embeddings}
For the RNN models, we make use of word features obtained from Word2Vec embeddings for the Hausa language and FastText embeddings for \yoruba and isiXhosa languages. We utilize the better quality embeddings recently released by \citet{Abdulmumin_2019} and \citet{alabi-LREC} for Hausa and \yoruba respectively instead of the pre-trained embeddings by Facebook that were trained on a smaller and lower quality dataset from Wikipedia. For isiXhosa, we did not find any existing word embeddings, therefore, we trained FastText embeddings from data collected from the I'solezwe\footnote{\url{https://www.isolezwelesixhosa.co.za/}} news website and the \textit{OPUS}\footnote{\url{http://opus.nlpl.eu/}} parallel translation website. The corpus size for isiXhosa is 1.4M sentences (around 15M tokens). We trained FastText embeddings for isiXhosa using \textit{Gensim}\footnote{\url{https://radimrehurek.com/gensim/}} with the following hyper-parameters: embedding size of 300, context window size of 5, minimum word count of 3, number of negative samples ten and number of iterations 10.  

\section{Distant Supervision}

\subsection{Distant supervision for Personal names, Organisation and Locations}
We make use of lists of entities to annotate PER, ORG and LOC automatically. In this paper, we extract personal names, organizations and locations from Wikidata as entity lists and assign a corresponding named entity label if tokens from an unlabeled text match an entry in an entity list. 

For NER, we use manual heuristics to improve matching. For \yoruba, a minimum token length of 3 was set for extraction of LOC and PER, while the minimum length for ORG was set to 2. This reduces the false positive rate, e.g. preventing matches with function words like ``of''. 

Applying this on the test set, we obtained a precision of $80\%$, $38\%$ and $28\%$ for LOC, ORG and PER respectively; a recall of $73\%$, $52\%$ and $14\%$ for LOC, ORG and PER respectively; and an F1-score of $76\%$, $44\%$ and $19\%$ for LOC, ORG and PER respectively. 

For Hausa NER, a minimum token length of 4 was set for extraction of LOC, ORG and PER. Based on these manual heuristics, on the test set, we obtained a precision of $67\%$, $12\%$ and $47\%$ for LOC, ORG and PER respectively; a recall of $63\%$, $37\%$ and $56\%$ for LOC, ORG and PER respectively; and an F1-score of $65\%$, $18\%$ and $51\%$ for LOC, ORG and PER respectively. 

\subsection{DATE rules for NER}
Rules allow us to apply the knowledge of domain experts without the manual effort of labeling each instance. We asked native speakers with knowledge of NLP to write DATE rules for Hausa and \yoruba. In both languages, date expressions are preceded by date keywords, like ``\textit{ranar}'' / ``\textit{\d oj\d{\'{o}}}'' (day), ``\textit{watan}'' / ``\textit{o\d{s}\`{u}}'' (month), and ``\textit{shekarar}'' / ``\textit{\d od{\d{\'{u}}}}n'' (year) in Hausa/\yoruba. For example, \textit{``18th of December, 2019''}  in Hausa / \yoruba translates to `` \textit{ranar 18 ga watan Disamba, shekarar 2019}'' / ``\textit{\d oj\d{\'{o}} 18 o\d{s}\`{u} \d{O}p{\d{\`{e}}}, \d{o}d\'{u}n 2019}''.  The annotation rules are based on these three criteria: (1) A token is a date keyword or follows a date keyword in a sequence. (2) A token is a digit, and (3) other heuristics to capture relative dates and date periods connected by conjunctions e.g between July 2019 and March 2020. Applying these rules result in a precision of $49.30\%/51.35\%$, a recall of $60.61\%/79.17\%$ and an F1-score of $54.42\%/62.30\%$ on Hausa /\yoruba test set respectively.  

\subsection{Rules for Topic classification}
For the \yoruba topic classification task, we collected terms that correspond to the different classes into sets. For example, the set for the class Nigeria contains names of agencies and organizations, states and cities in Nigeria. The set for the World class is made up of the name of countries of the world, their capitals and major cities and world affairs related organizations. Names of sporting clubs and sportspeople across the world were used for the Sports class and list of artists and actresses and entertainment-related terms for the Entertainment class. Given a news headline to be annotated, we get the union set of 1{-} and 2{-}grams and obtain the intersection with the class dictionaries we have. The class with the highest number of intersecting elements is selected. In the case of a tie, we randomly pick a class out the classes with a tie. 
Just as we did for  \yoruba, we collected the class-related tokens for the Hausa text classification. We, however, split the classification into two steps, checking some important tokens and using the same approach as we used for \yoruba. If a headline contains the word \textit{cutar} (disease) , it is classified as Health, if it contains tokens such as \textit{inec}, \textit{zaben}, \textit{pdp},\textit{apc} (which are all politics related tokens) it is classified as Politics. Furthermore, sentences with any of the following tokens \textit{buhari}, \textit{legas}, \textit{kano}, \textit{kaduna}, \textit{sokoto} are classified as Nigeria while sentences with \textit{afurka}, \textit{kamaru} and \textit{nijar} are classified as Africa. If none of the tokens highlighted above is found, we apply the same approach as we did for the  \yoruba setting, which is majority voting of the intersection set of the news headline with a keyword set for each class.
Applying these rules results in a precision of $59.54\%/60.05\%$, a recall of $46.04\%/53.66\%$ and an F1-score of $48.52\%/54.93\%$ on the Hausa /\yoruba test set respectively. 

\section{Experimental Settings}

\begin{figure*}
\centering
    \begin{subfigure}[b]{0.32\textwidth}
        \centering
        \includegraphics[height=4cm]{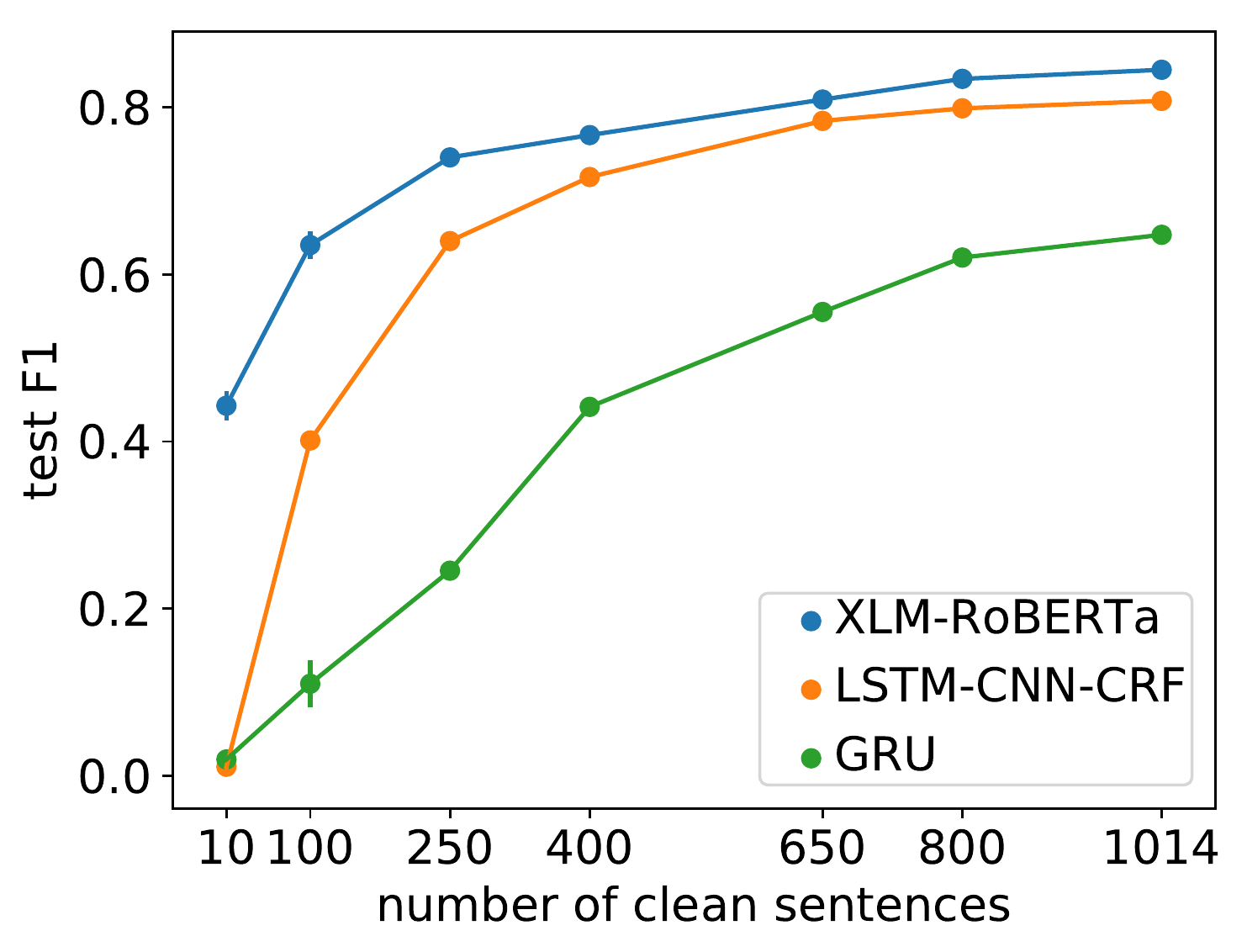}
        \caption{NER Hausa}
    \end{subfigure}
    \begin{subfigure}[b]{0.32\textwidth}
        \centering
        \includegraphics[height=4cm]{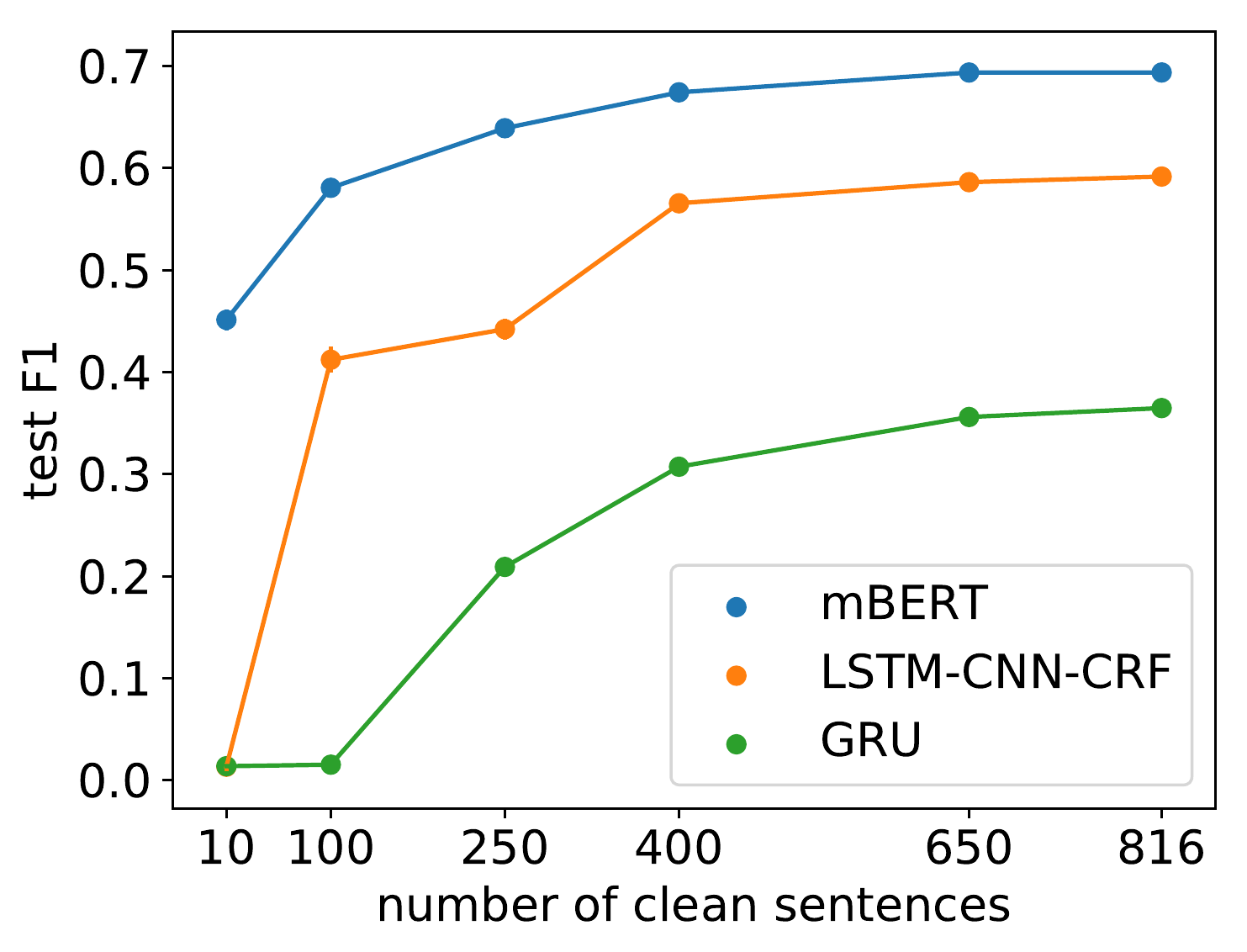}
        \caption{NER \yoruba}
    \end{subfigure}
    \begin{subfigure}[b]{0.32\textwidth}
        \centering
        \includegraphics[height=4cm]{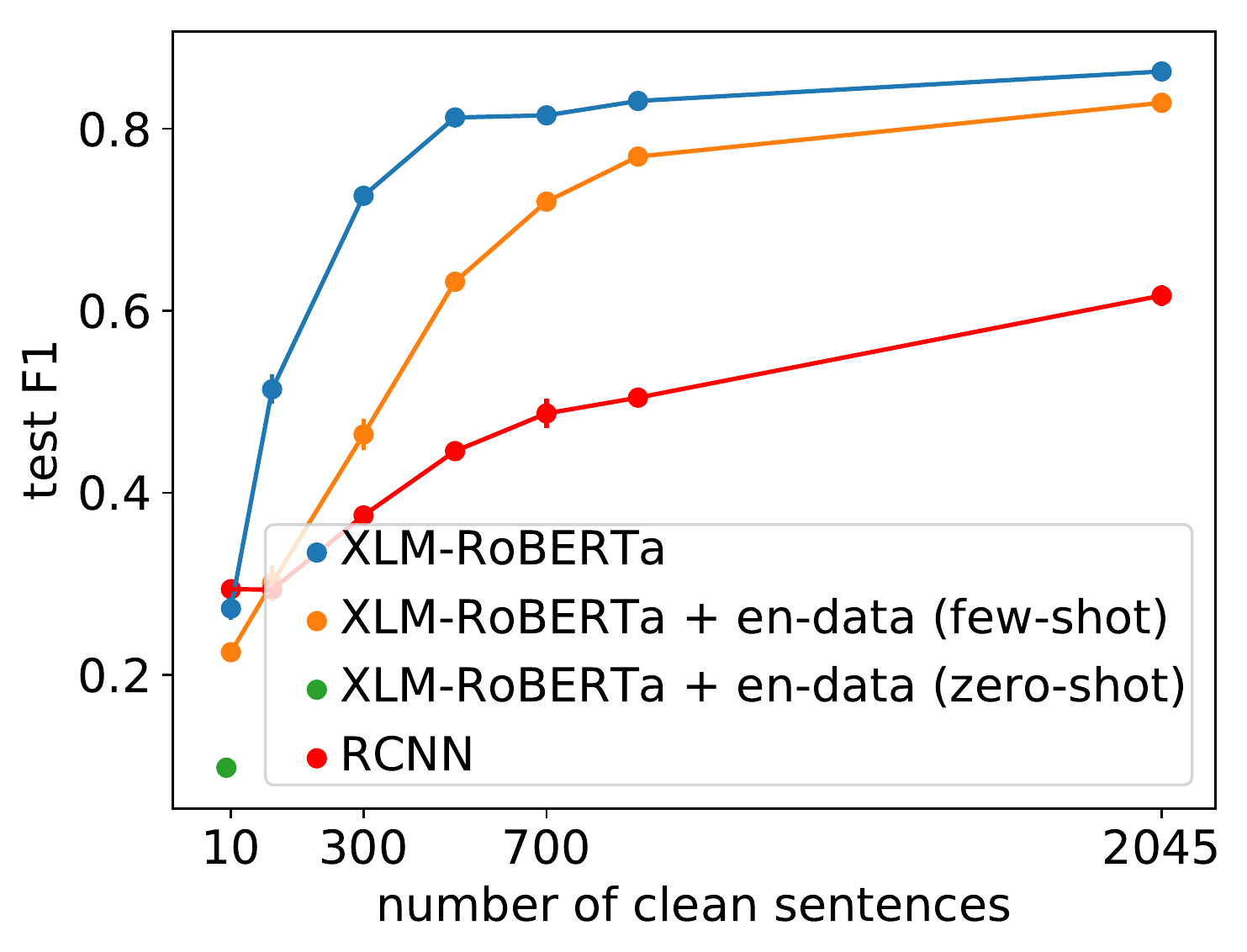}
        \caption{Topic Class. Hausa}
    \end{subfigure}
    \begin{subfigure}[b]{0.32\textwidth}
     \centering
        \includegraphics[height=4cm]{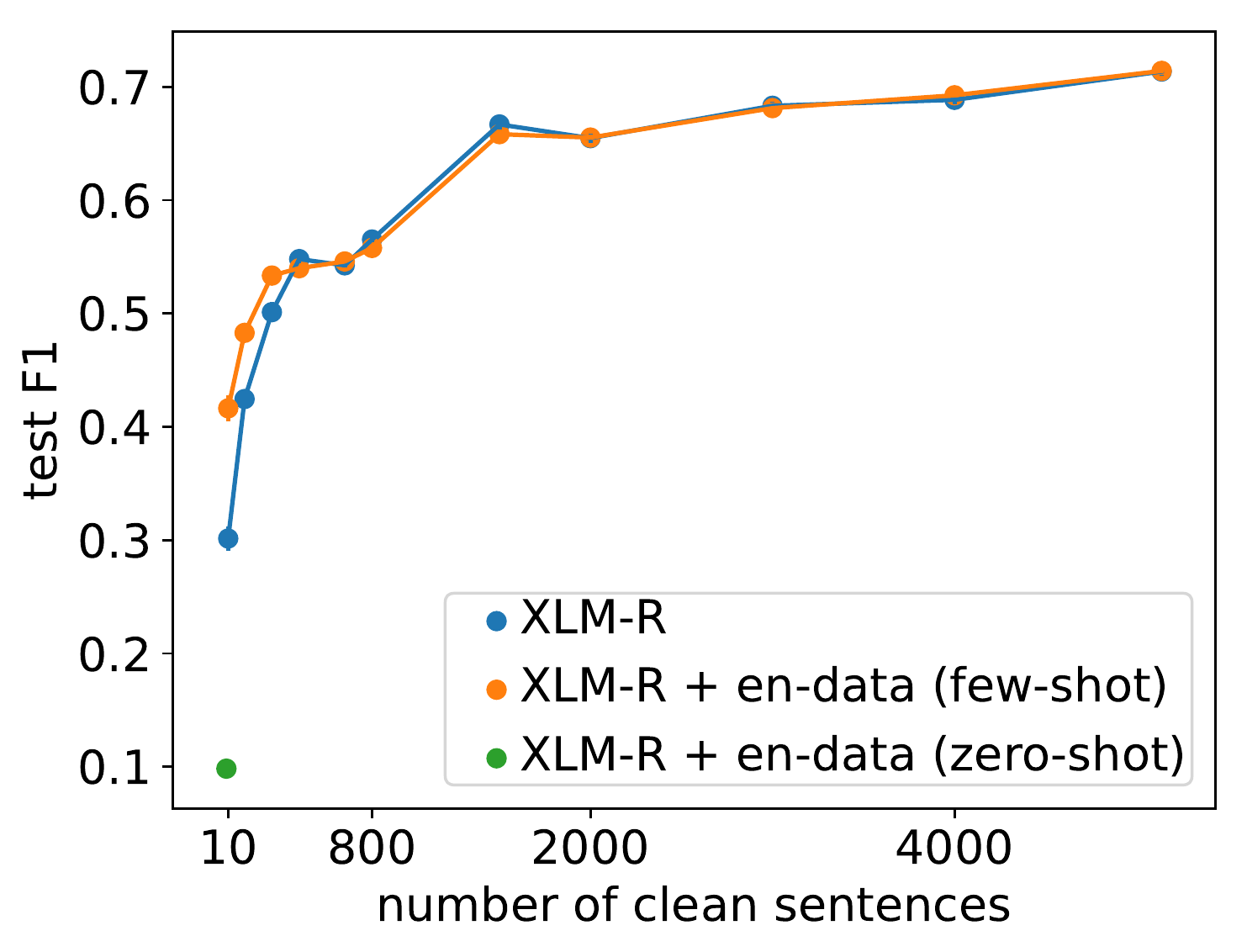}
        \caption{Transfer Learn NER isiXhosa}
    \end{subfigure}
    \begin{subfigure}[b]{0.33\textwidth}
        \centering
        \includegraphics[height=4cm]{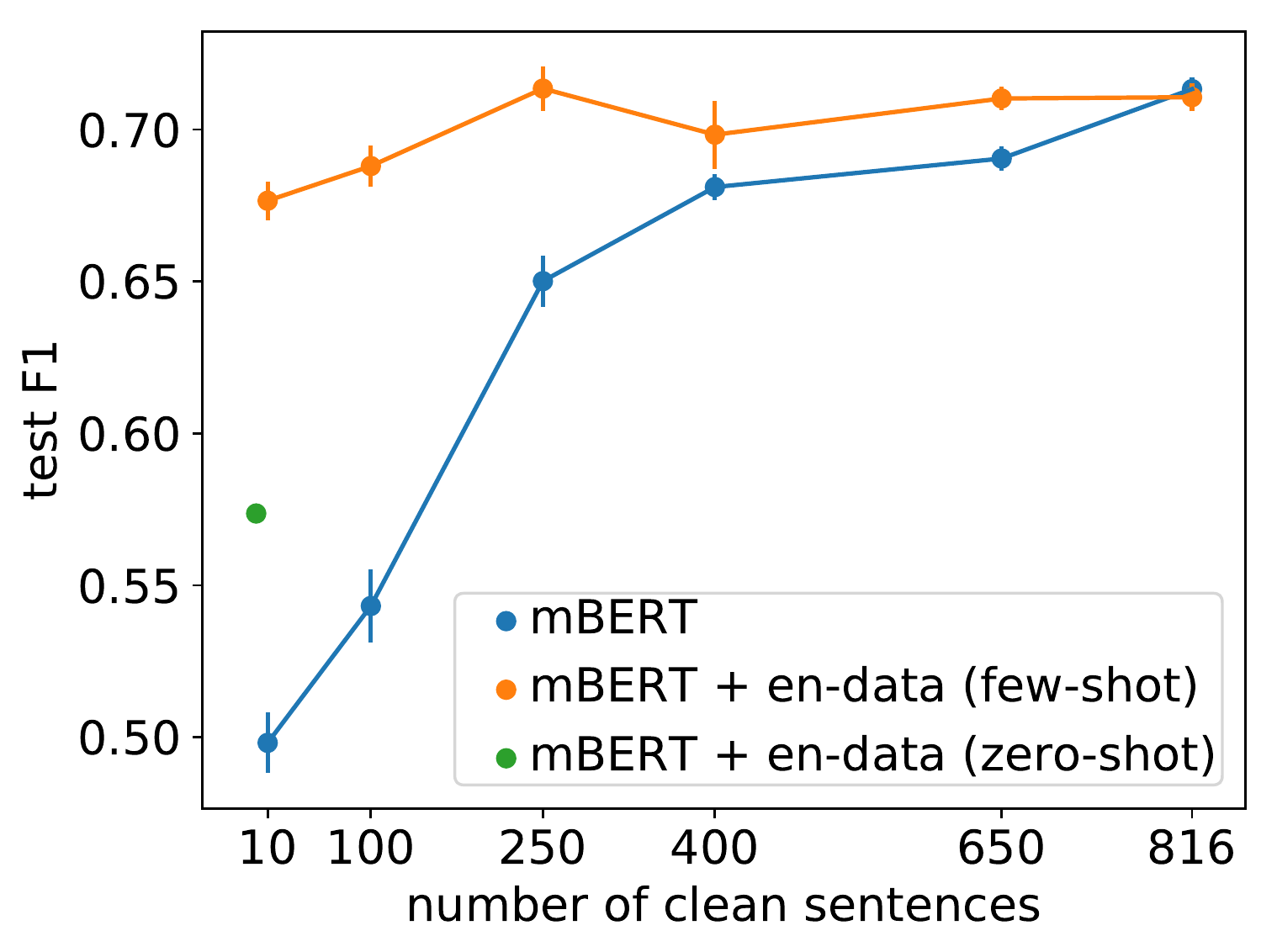}
        \caption{Transfer Learn NER \yoruba}
    \end{subfigure}
    \begin{subfigure}[b]{0.32\textwidth}
        \centering
        \includegraphics[height=4cm]{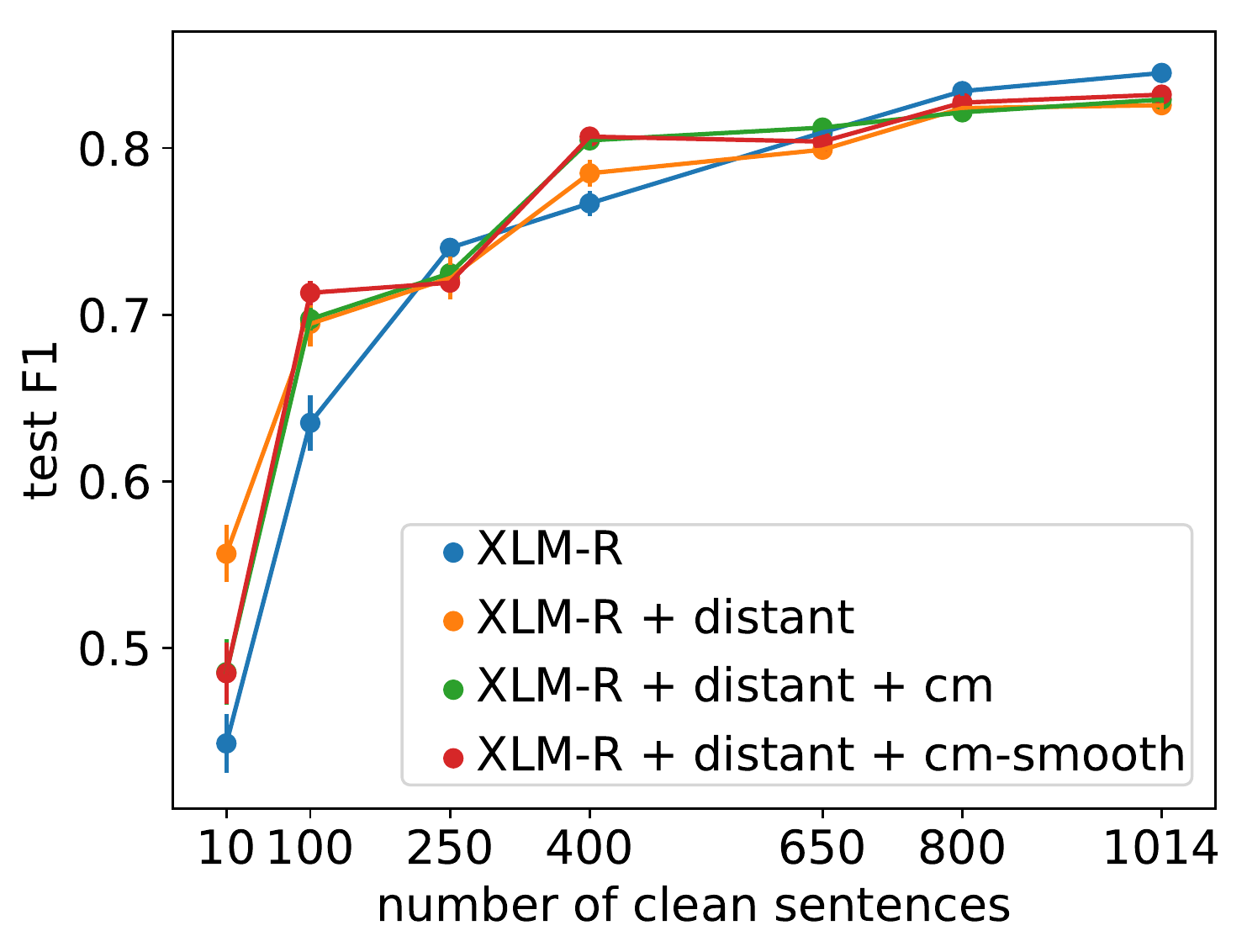}
        \caption{Distant NER Hausa}
    \end{subfigure}
    \begin{subfigure}[b]{0.32\textwidth}
        \centering
        \includegraphics[height=4cm]{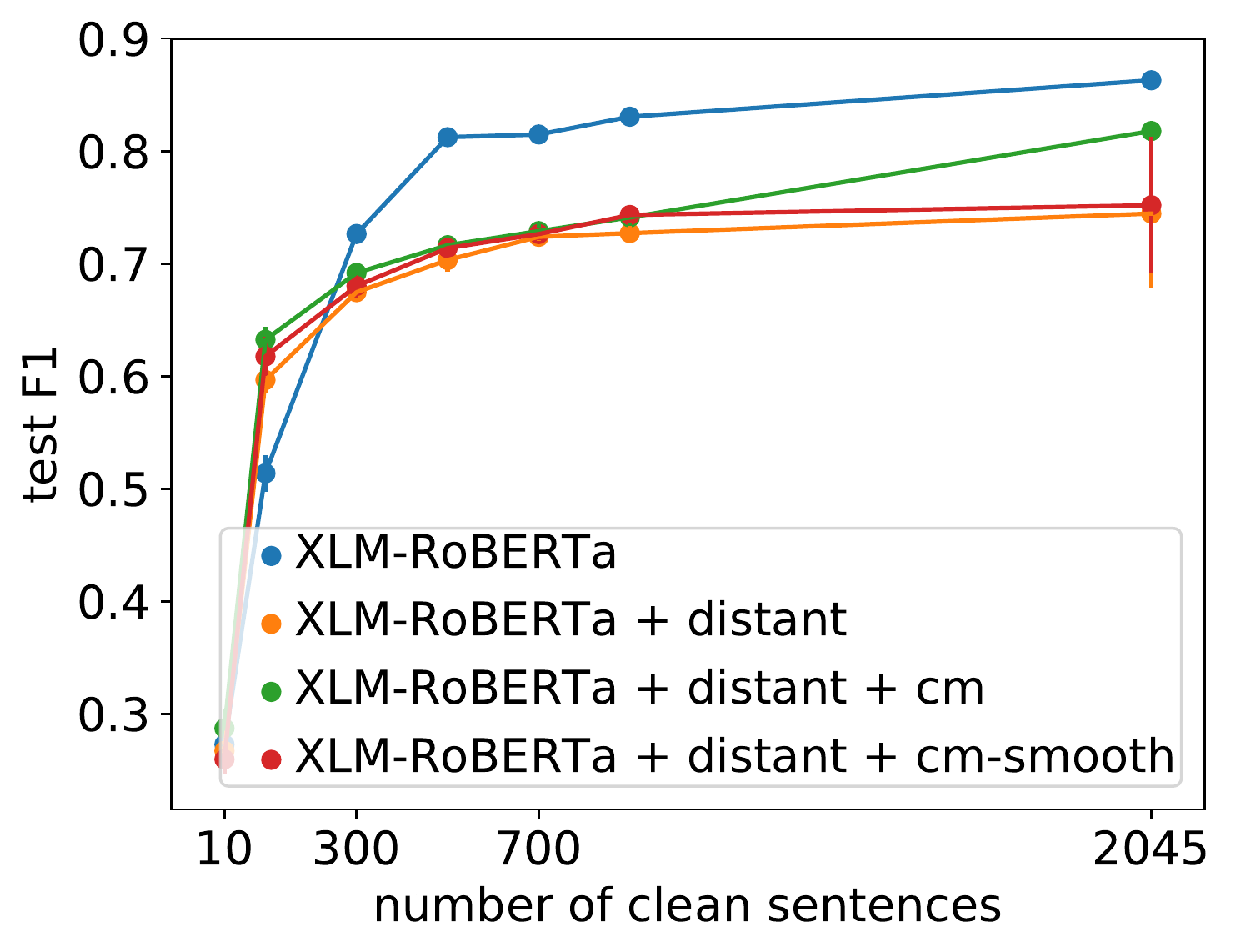}
        \caption{Distant Topic Class. Hausa}
    \end{subfigure}
    \begin{subfigure}[b]{0.32\textwidth}
        \centering
        \includegraphics[height=4cm]{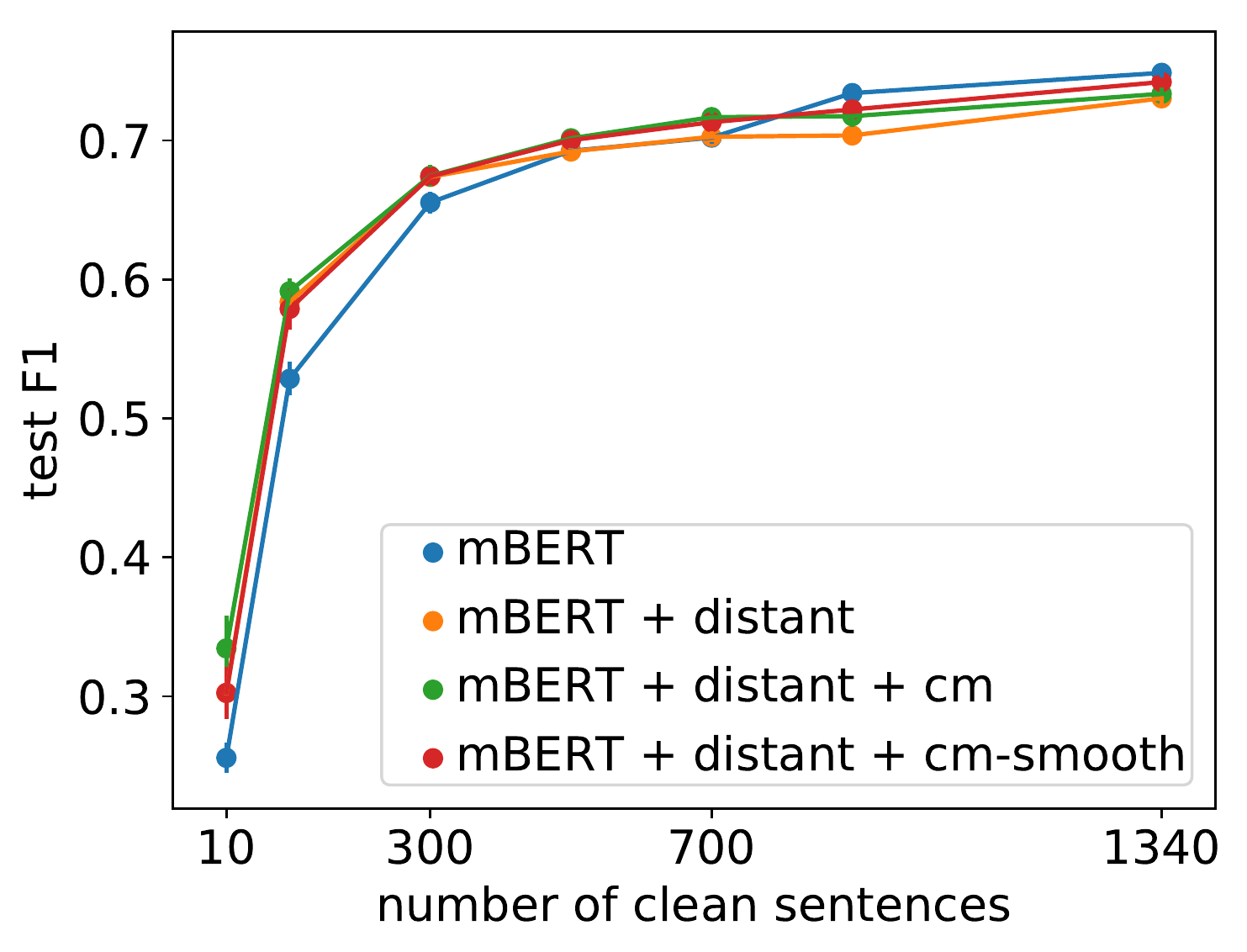}
        \caption{Distant Topic Class. \yoruba}
    \end{subfigure}
    \begin{subfigure}[b]{0.32\textwidth}
        \centering
        \includegraphics[height=4cm]{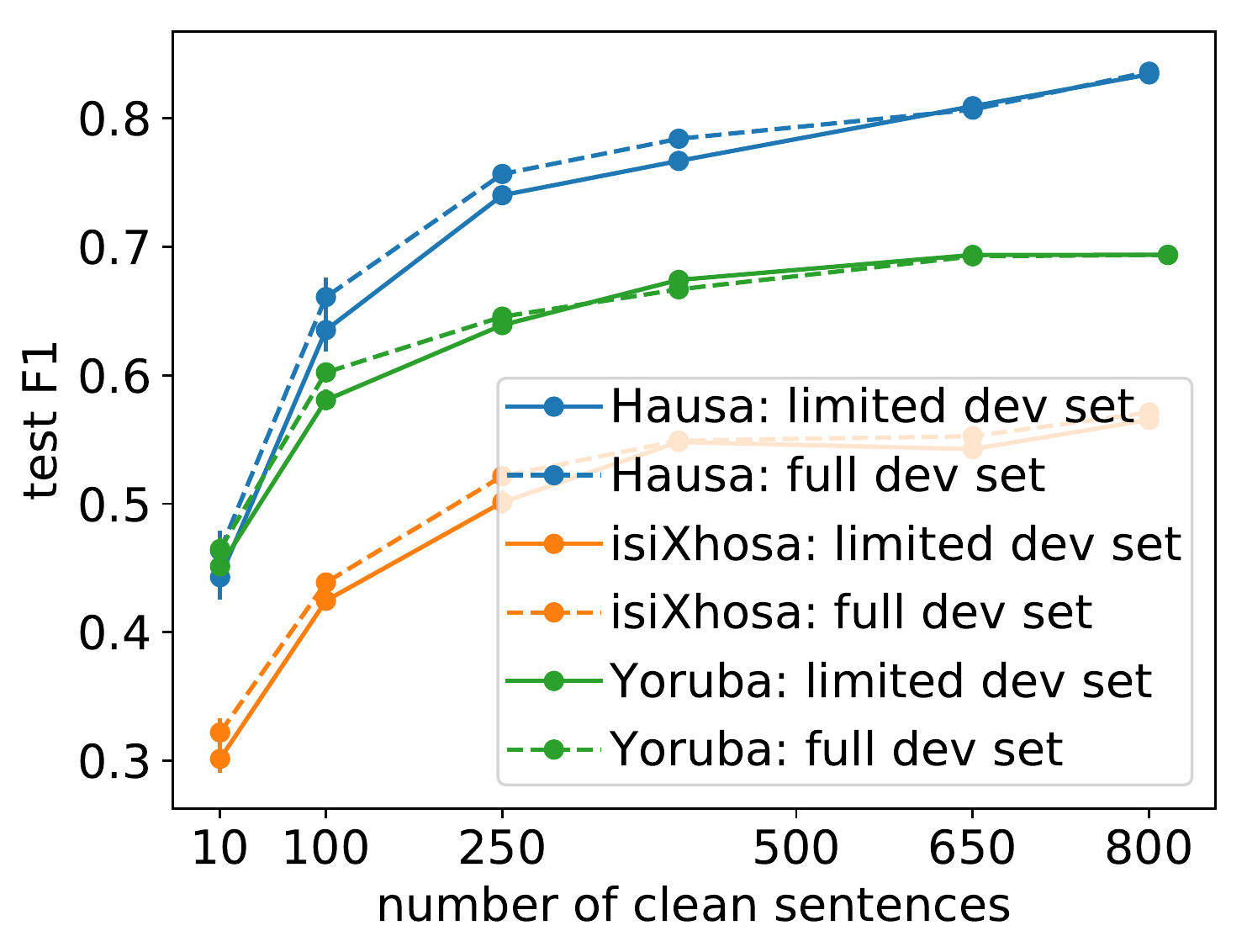}
        \caption{Development Set NER}
    \end{subfigure}
    \caption{Additional plots.\label{fig:add_plots}}
\end{figure*}

\subsection{General}
All experiments were repeated ten times with varying random seeds but with the same data (subsets). We report mean F1 test score and standard error ($\sigma / \sqrt{10}$). For NER, the score was computed following the standard CoNLL approach \cite{data/CoNLL03} using the \textit{seqeval} implementation.\footnote{\url{https://github.com/chakki-works/seqeval}} Labels are in the BIO2-scheme. For evaluating topic classification, the implementation by \textit{scikit-learn} was used.\footnote{\url{https://scikit-learn.org/stable/modules/generated/sklearn.metrics.classification_report.html}} All models are trained for 50 epochs, and the model that performed best on the (possibly size-reduced) development set is used for evaluation. 

\subsection{BERT and XLM-RoBERTa}

As multilingual transformer models, mBert and XLM-RoBERTa are used, both in the implementation by \citet{wolf2019huggingfaces}. The specific model IDs are \textit{bert-base-multilingual-cased} and \textit{xlm-roberta-base}.\footnote{\url{https://huggingface.co/transformers/pretrained_models.html}} For the DistilBERT experiment it is \textit{distilbert-base-multilingual-cased}. As is standard, the last layer (language model head) is replaced with a classification layer (either for sequence or token classification). Models were trained with the Adam optimizer and a learning rate of $5e^{-5}$. Gradient clipping of value 1 is applied. The batch size is 32 for NER and 128 for topic classification.
For distant supervision and XLM-RoBERTa on the Hausa topic classification data with 100 or more labeled sentences, we observed convergence issues where the trained model would just predict the majority classes. We, therefore, excluded for this task runs where \textit{on the development set} the class-specific F1 score was 0.0 for two or more classes. The experiments were then repeated with a different seed.  

\subsection{Other Architectures}

For the GRU and LSTM-CNN-CRF model, we use the implementation by \citet{chernodub2019targer} with modifications to support FastText embeddings and the \textit{seqeval} evaluation library. Both model architectures are bidirectional. Dropout of 0.5 is applied. The batch-size is 10 and SGD with a learning rate of 0.01, and a decay of 0.05 and momentum of 0.9 is used. Gradients are clipped with a value of 5. The RNN dimension is 300. For the CNN, the character embedding dimension is 25 with 30 filters and a window-size of 3.

For the topic classification task, we experiment with the RCNN model proposed by \cite{lai2015recurrent}. The hidden size in the Bi-LSTM is 100 for each direction. The linear layer after the Bi-LSTM reduces the dimension to 64. The model is trained for 50 epochs.

\subsection{Transfer Learning}

For transfer learning, the model is first fine-tuned on labeled data from a high-resource language. Following \cite{data/Xtreme20}, we use the English CoNLL03 NER dataset \cite{data/CoNLL03} for NER. It consists of ca. 8k training sentences. The model is trained for 50 epochs and the weights of the best epoch according to the development set are taken. The training parameters are the same as before. On the English CoNLL03 test set, the model achieves a performance of 0.90 F1-score. As the Hausa and \yoruba datasets have slightly different label sets, we only use their intersection, resulting in the labels PER, LOC and ORG and excluding MISC from CoNLL03 and the DATE label from Hausa/\yoruba. For isiXhosa, the label sets are identical (i.e. also including MISC). After fine-tuning the model on the high-resource data, the model is directly evaluated on the African test set (for zero-shot) or fine-tuned and then evaluated on the African data (for few-shot).

For topic classification, the AG News corpus is used \cite{data/ag-news}. It consists of 120k training sentences. The model is trained for 20 epochs and the weights of the best epoch according to the test set are used. On this set, an F1-score of 0.93 is achieved. The training procedure is the same as above. For the labels, we use the union of the labels of the AG News corpus (Sports, World, Business and Sci/Tech) and the African datasets.

\subsection{Label Noise Handling}

We use a confusion matrix which is a common approach for handling noisy labels (see, e.g. \cite{fang-cohn-2016-learning, luo-etal-2017-learning-noise, lange-etal-2019-feature, wang-etal-2019-learning-noisy}). The confusion matrix models the relationship between the true, clean label of an instance and its corresponding noisy label. When training on noisy instances, the confusion matrix is added to the output of the main model (that usually predicts clean labels) changing the output label distribution from the clean to the noisy one. This allows to then train on noisily labeled instances without a detrimental loss obtained by predicting the true, clean label but having noisy, incorrect labels as targets.

We use the specific approach by \citet{noise/Hedderich18Confusion} that was developed to work with small amounts of manually labeled, clean data and a large amount of automatically annotated, noisy labels obtained through distant supervision. To get the confusion matrix of the noise, the distant supervision is applied on the small set of clean training instances. From the resulting pairs of clean and noisy labels, the confusion matrix can be estimated. 

In a setting where only a few instances are available, the estimated confusion matrix might not be close to the actual change in the noise distribution. We, therefore, combine it with the smoothing approach by \citet{noise/LvSmoothing20}. Each entry of the probabilistic confusion matrix is raised to the power of $\beta$ and then row-wise normalized.

As studied by \citet{noise/Hedderich18Confusion}, we do not use the full amount of available, distantly supervised instances in each epoch. Instead, in each epoch, only a randomly selected subset of the size of the clean, manually labeled training data is used to lessen the negative effects of the noisy labels additionally. For smoothing, $\beta=0.8$ is used as this performed best for \citet{noise/LvSmoothing20}.

\end{document}